%% file: main.tex
\documentclass[10pt,twocolumn,letterpaper]{article} 
\usepackage[pagenumbers]{cvpr} 
\input{math_commands.tex}

\usepackage{hyperref}
\usepackage{url}
\usepackage{caption}
\usepackage{graphicx}
\usepackage{amsmath}
\usepackage{bbm}
\usepackage{amsthm}
\usepackage{booktabs}
\usepackage{algorithm}
\usepackage{algpseudocode}
\usepackage{alltt}
\usepackage{amssymb}
\usepackage{multirow}
\usepackage{svg}
\usepackage{wrapfig}
\usepackage{tikz}
\usepackage{cancel} 
\usepackage{makecell}
\usepackage{colortbl}
\usepackage{bbding}
\usepackage{color,xcolor}
\usepackage{bbm}
\usepackage{pifont} 
\newcommand{\cmark}{\ding{51}}%
\newcommand{\xmark}{\ding{55}}%

\definecolor{C0}{rgb}{0.121569, 0.466667, 0.705882}
\definecolor{C1}{rgb}{1.000000, 0.498039, 0.054902}
\definecolor{C2}{rgb}{0.172549, 0.627451, 0.172549}
\definecolor{C3}{rgb}{0.839216, 0.152941, 0.156863}
\definecolor{C4}{rgb}{0.580392, 0.403922, 0.741176}
\definecolor{C5}{rgb}{0.549020, 0.337255, 0.294118}
\definecolor{C6}{rgb}{0.890196, 0.466667, 0.760784}
\definecolor{C7}{rgb}{0.498039, 0.498039, 0.498039}
\definecolor{C8}{rgb}{0.737255, 0.741176, 0.133333}
\definecolor{C9}{rgb}{0.090196, 0.745098, 0.811765}

\newcommand\RR[1]{{\color{C1}#1}}

\newcommand\GG[1]{{\color{C2}#1}}

\definecolor{aliceblue}{RGB}{255, 245, 165}
\newcommand{\CC}{\cellcolor{aliceblue}}
\definecolor{babyblue}{RGB}{255, 255, 255}
\newcommand{\CB}{\cellcolor{babyblue}}

\newcommand\Add[1]{{\color{C6} ($+\Delta$ #1)}}
\newtheorem{theorem}{Theorem}[section]

\usepackage{setspace}
\usepackage{algorithmicx}

\definecolor{Blue}{RGB}{0, 183, 133}
\definecolor{Aquamarine}{RGB}{127, 255, 212}
\definecolor{Sepia}{RGB}{112, 66, 20}
\definecolor{BrickRed}{RGB}{203, 65, 84}
\definecolor{ForestGreen}{RGB}{34,139,34}
\definecolor{Black}{RGB}{0,0,0}
\colorlet{my-red}{BrickRed!90!Sepia}
\colorlet{my-kk}{Sepia!30!Blue}
\colorlet{my-blue}{Aquamarine!30!Blue}
\hypersetup{
  colorlinks,
  urlcolor  = my-red,
  linkcolor = my-kk,
  citecolor = my-blue,
}
\usepackage{tikz}

\newlength{\originaljot}
\usepackage[most]{tcolorbox}

\tcbset{
  shaded citation/.style={
    colback=gray!20, 
    colframe=gray!30, 
    sharp corners, 
    boxrule=0.5pt, 
    drop shadow={gray!50!white}, 
    left=0.5mm, right=0.5mm, top=0.5mm, bottom=0.5mm, 
    enhanced,
  }
}
\newcommand{\imineq}[2]{\vcenter{\hbox{\includegraphics[height=#2ex]{#1}}}}

\newcommand{\IVS}{\textit{IV-mixed Sampler}}
\newcommand{\IDM}{\epsilon_\theta^{\tiny \mathbf{\mathrm{I}}}(\cdot,\cdot)}
\newcommand{\VDM}{\epsilon_\theta^{\tiny \mathbf{\mathrm{V}}}(\cdot,\cdot)}
\usepackage{tikz}
\usepackage{mathtools}
\usepackage[normalem]{ulem}  %
\newcommand\model{\bm{\epsilon}_\theta}
\newcommand\ve[1]{\mathbf{#1}}

\title{\textit{IV-Mixed Sampler}: Leveraging Image Diffusion Models\\ for Enhanced Video Synthesis}
\author{Shitong Shao$^1$ \qquad Zikai Zhou$^{1,\diamondsuit}$ \qquad Lichen Bai$^{1}$ \qquad Haoyi Xiong$^{2}$ \qquad Zeke Xie$^{1,*}$ \\
  $^1$Hong Kong University of Science and Technology (Guangzhou)  $^2$Baidu Inc. \\
{\tt\small \{sshao213,zikaizhou,lichenbai,zekexie\}@hkust-gz.edu.cn} \\
{\tt\small haoyi.xiong.fr@ieee.org,\ $^\diamondsuit$:Equal Contribution, $*$:Corresponding author}}

\begin{document}
\twocolumn[{%
\renewcommand\twocolumn[1][]{#1}%
\maketitle\begin{center}\centering\includegraphics[width=1.0\linewidth]{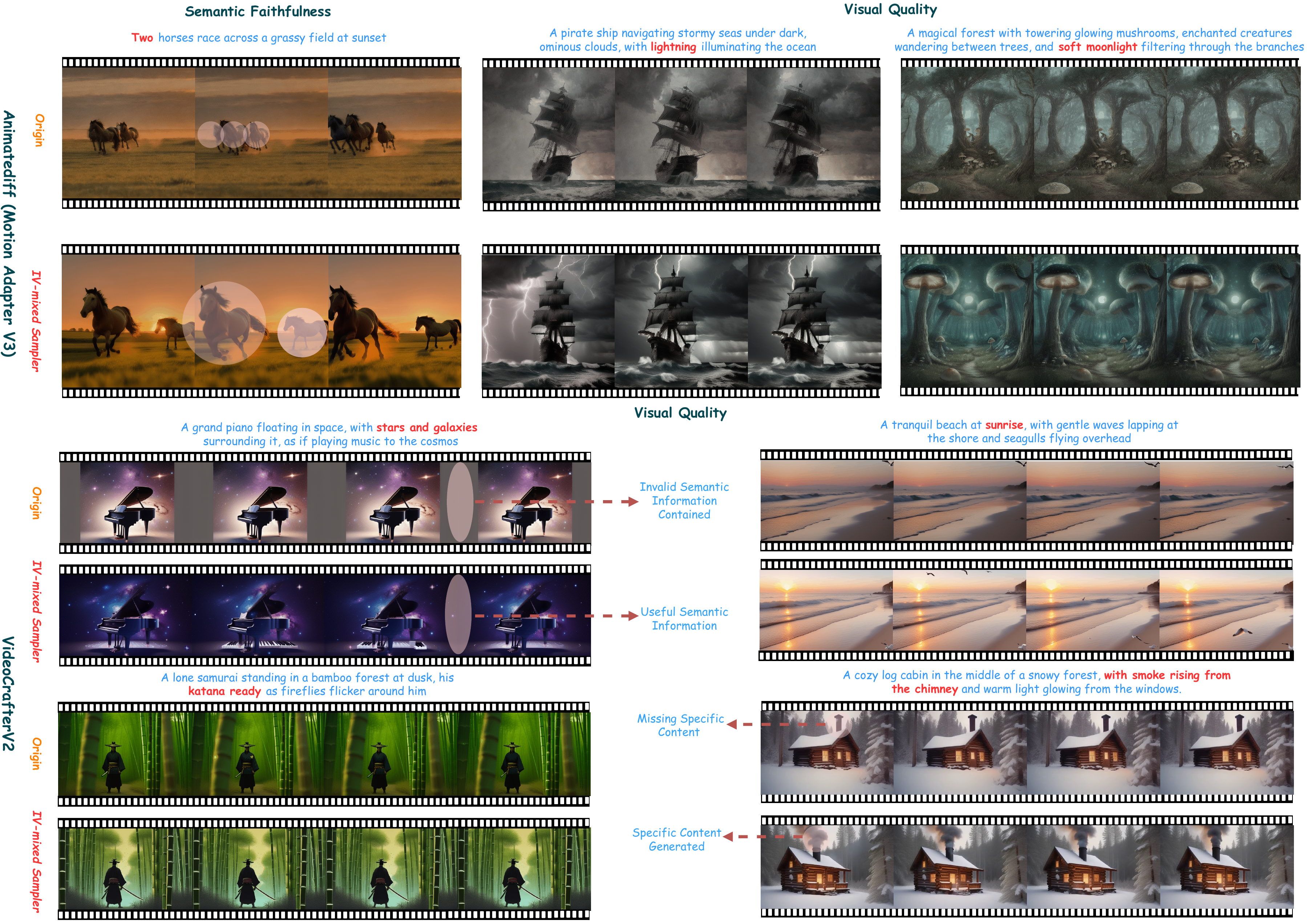}
\captionof{figure}{\textbf{Visualization of \IVS\ and the standard DDIM sampling on Animatediff and VideoCrafterV2.} Unlike prior heavy-inference approaches~\citep{guo2024i4vgen,freeinit}, \IVS\ is able to significantly improve the fidelity of the video while guaranteeing semantic faithfulness.}
\label{figure:visualization_main_paper}
\end{center}%
}]

\maketitle

\begin{abstract}
The multi-step sampling mechanism, a key feature of visual diffusion models, has significant potential to replicate the success of OpenAI's Strawberry in enhancing performance by increasing the inference computational cost. Sufficient prior studies have demonstrated that correctly scaling up computation in the sampling process can successfully lead to improved generation quality, enhanced image editing, and compositional generalization. While there have been rapid advancements in developing inference-heavy algorithms for improved image generation, relatively little work has explored inference scaling laws in video diffusion models (VDMs). Furthermore, existing research shows only minimal performance gains that are perceptible to the naked eye. To address this, we design a novel training-free algorithm \textit{IV-Mixed Sampler} that leverages the strengths of image diffusion models (IDMs) to assist VDMs surpass their current capabilities. The core of \textit{IV-Mixed Sampler} is to use IDMs to significantly enhance the quality of each video frame and VDMs ensure the temporal coherence of the video during the sampling process. Our experiments have demonstrated that \textit{IV-Mixed Sampler} achieves state-of-the-art performance on 4 benchmarks including UCF-101-FVD, MSR-VTT-FVD, Chronomagic-Bench-150, and Chronomagic-Bench-1649. For example, the open-source Animatediff with \textit{IV-Mixed Sampler} reduces the UMT-FVD score from 275.2 to 228.6, closing to 223.1 from the closed-source Pika-2.0.

\end{abstract}

\section{Introduction}
\label{sec:introduction}
In the large foundation models era, maximizing the inference potential of foundation models~\citep{blattmann2023stable,llama} that require high pre-training costs has become a research staple for academics~\citep{freeinit,CoT}. Efficient plug-and-play algorithms~\citep{controlnet,mou2024t2i} can significantly drive large-scale models to reach their full potential and outperform the original counterparts due to low trial-and-error costs. In contrast to popular inference-heavy algorithms (\textit{e.g.}, chain-of-thought (CoT)~\citep{CoT} and OpenAI's Strawberry~\citep{strawberry}), which consistently emerge in the large language model (LLM) field, text-to-video (T2V) synthesis still faces the prevalent challenge of low-quality synthesized videos that lack semantic faithfulness~\citep{animatediff, modelscope}. This limitation severely hampers the deployment and application of video diffusion models (VDMs). Motivated by the success of inference scaling laws in LLMs, we inevitably wonder whether a Markovian chain exists within VDMs similar to that of the autoregressive models~\citep{transformer,devlin2018bert}, allowing us to utilize its properties to enhance VDMs' inference performance? The answer is obvious because the diffusion model is established on differential equations that incorporate both a forward and a reverse process~\citep{sde}, allowing it to be fully considered a Markov chain during inference.
\begin{figure}[!h]
\vspace{-0.2in}
\includegraphics[height=0.42\textwidth,trim={0cm 0cm 0cm 0cm},clip]{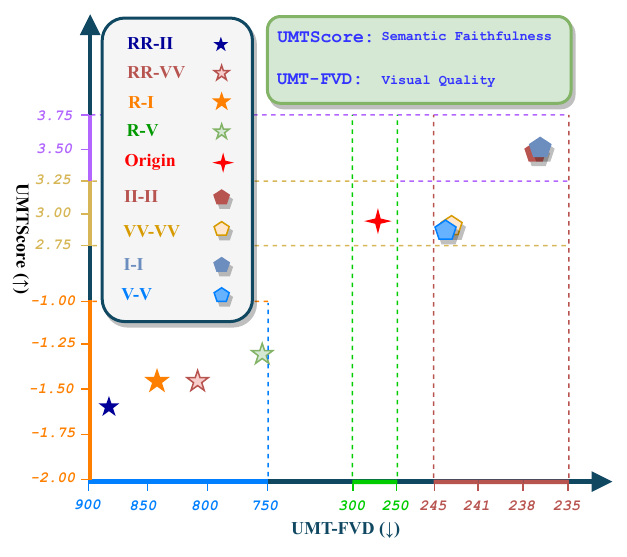}
\vspace{-0.1in}
\caption{\small UMTScore (↑) vs. UMT-FVD (↓) with Animatediff~\citep{animatediff} on Chronomagic-Bench-150~\citep{yuan2024chronomagic}. In the legend, ``R'', ``I'', and ``V'' represent the score function estimation using random Gaussian noise, IDM, and VDM, respectively. Moreover, the front of the horizontal line ``-'' refers to the additive noise form, while the back of ``-'' represents the denoising paradigm. For instance, ``RR-II'' stands for a two-step of adding noise with Guassian noise followed by two-step of denoising performed using IDM.}
\label{figure:motivation}
\end{figure}
\paragraph{Motivation.} VDM-based training-free algorithms typically face a significant challenge: their performance ceiling is constrained by the VDM itself. To be specific, the videos synthesized from most open-source VDMs~\citep{animatediff,modelscope} exhibit several inherent problems, such as weak semantic consistency between the videos and the prompts, as well as low quality. A widely accepted perspective~\citep{freeinit,mcm_accelerate} on this phenomenon is that underperforming VDMs cannot overcome their intrinsic limitations because they are trained on low-quality, small-scale T2V datasets. In contrast, image diffusion models (IDMs) can now reliably serve commercial application scenarios~\citep{della3}, thanks to the well-established dataset ecosystem created by the AIGC community. This discrepancy naturally prompts us to consider whether we can view IDMs as tools to enhance VDMs and how to make IDMs effective assistants for VDMs.


Since diffusion models employ a multi-step sampling mechanism and the quality of synthesized samples from IDMs is significantly higher than that from VDMs, it is reasonable to use IDMs and VDMs alternately for score function estimation throughout the reverse sampling process. However, this straightforward paradigm leads to a loss of temporal coherence in the synthesized videos during our initial empirical explorations. Inspired by SDEdit~\citep{meng2021sdedit}, we enhance the quality of each frame at every denoising step by performing the following additional operation: \textit{\textcolor{C3}{1)} first adding Gaussian noise and \textcolor{C3}{2)} then denoising using IDMs.} Unfortunately, as illustrated in Fig.~\ref{figure:motivation}, the approach ``R-$[\cdot]$'', which use Gaussian noise to perform the forward diffusion process, result in significantly lower quality of the synthesized video compared to the standard DDIM process (\textit{i.e.}, Origin in Fig.~\ref{figure:motivation}). This phenomenon arises because ``R-$[\cdot]$'' over-introduces invalid information (\textit{i.e.}, Gaussian noise) into the synthesized video during denoising. Given this, we consider the more robust \textit{deterministic sampling} method to integrate the video denoising process with the image denoising process, as this paradigm is stable and effectively reduces truncation errors in practical discrete sampling.

In this paper, we consider the widely used \textit{deterministic function} DDIM-Inversion~\citep{ddiminversion} to inject perturbations into the video. The experimental results of labeled ``I-$[\cdot]$'' and ``V-$[\cdot]$'' in Fig.~\ref{figure:motivation} demonstrate that first performing DDIM-Inversion and then executing DDIM before each denoising step significantly improves video quality. Additionally, the observation that ``I-$[\cdot]$'' significantly surpasses ``V-$[\cdot]$'' further substantiates that IDMs trained on high-quality datasets can play a positive role in the video sampling process. Motivated by this, we further explore the upper bound of the performance gain that IDM provides for video synthesis. We primarily extend the paradigm of the single-step diffusion process and the single-step reverse process to multiple steps. Then, to ensure inference efficiency, we investigate all possible combinations in the two-step diffusion process and the two-step inverse process, collectively naming this series of algorithms \IVS.

\paragraph{Contribution.} Specifically, \textcolor{C3}{\textit{1)}} we construct \IVS\ under a rigorous mathematical framework and demonstrate, through theoretical analysis, that it can be elegantly transformed into a standard inverse ordinary differential equation (ODE) process. For the sake of intuition, we present \IVS\ (\textit{w.r.t.}, ``IV-IV'') on Fig.~\ref{figure:illustration} and its pseudo code in Appendix~\ref{apd:pseudo_code}. \textcolor{C3}{\textit{2)}} The empirically optimal \IVS\ further reduces the UMT-FVD by approximately 10 points compared to the best ``I-I'' in Fig.~\ref{figure:motivation} and by 39.72 points over FreeInit~\citep{freeinit}. Furthermore, \textcolor{C3}{\textit{3)}} we conduct sufficient ablation studies to determine which classifier-free guidance (CFG)~\citep{nips2021_classifier_free_guidance} scale and which sampling paradigm yield the best performance for various metrics at what sampling intervals. In addition to this, \textcolor{C3}{\textit{4)}} qualitative and quantitative comparison experiments have amply demonstrated that our algorithm achieves state-of-the-art (SOTA) performance on four popular benchmarks: UCF-101-FVD~\citep{soomro2012ucf101}, MSR-VTT-FVD~\citep{msrvtt}, Chronomagic-Bench-150~\citep{yuan2024chronomagic}, and Chronomagic-1649~\citep{yuan2024chronomagic}. These competitive outcomes demonstrate that our proposed \IVS\ dramatically improves the visual quality and semantic faithfulness of the synthesized video. 

\section{Preliminary}

 We review VDMs, SDEdit, DDIM \& DDIM-Inversion in this section and further bootstrap how past work has designed VDM-based plug-and-play algorithms in Appendix~\ref{sec:heavy_inference}.

\begin{figure*}[t]
\centering
\includegraphics[width=0.95\textwidth,trim={0cm 0cm 0cm 0cm},clip]{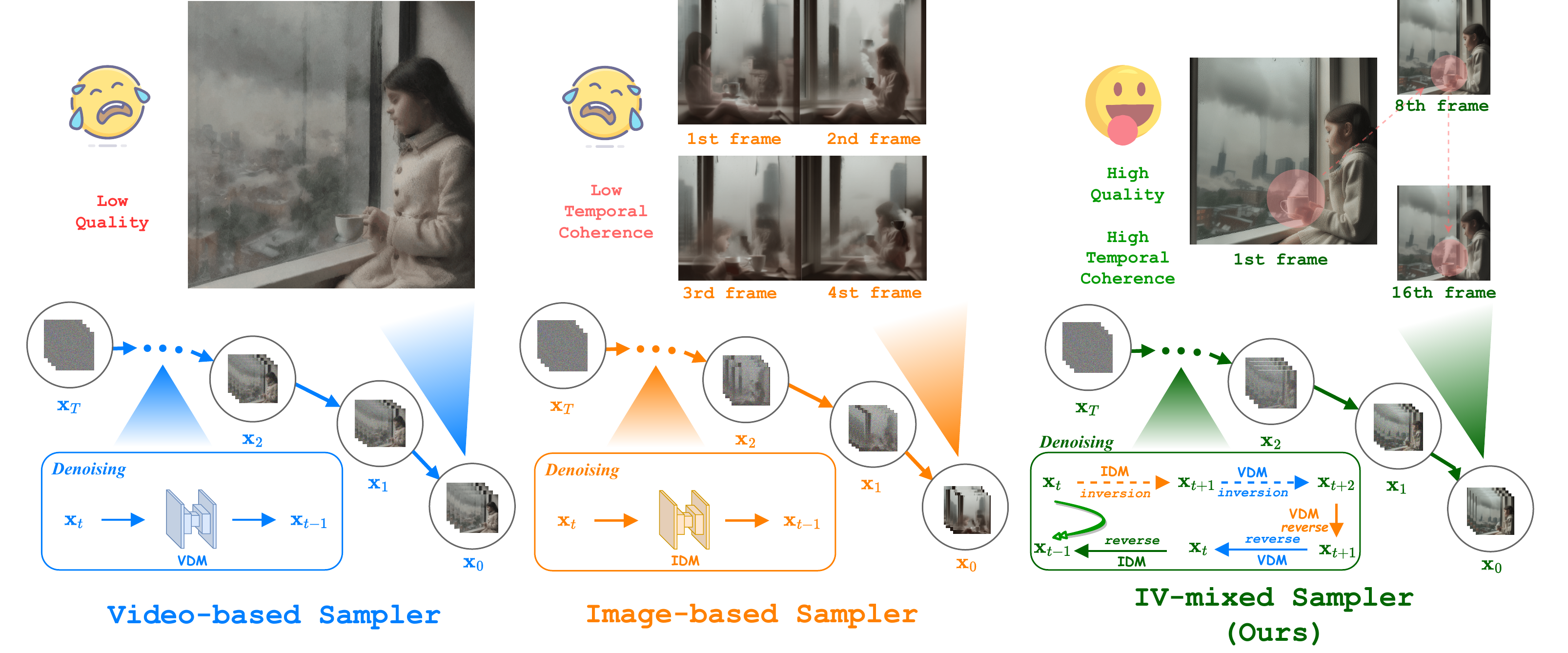}
\vspace{-2.2ex}
\caption{\textbf{Overview of our \textit{IV-mixed Sampler}, Video-based Sampler and Image-based Sampler.} \IVS\ utilizes IDM and VDM to ensure synthesized video quality and temporal coherence, respectively.}
\vspace{-3ex}
\label{figure:illustration}
\end{figure*}

\paragraph{Diffusion Models.} Diffusion models~\citep{ddpm_begin,sde,ddim} including IDMs and VDMs consist of a forward process and a reverse process. Given $\ve{x}_0$ represent a \textit{D}-dimensional random variable sampled from the real data distribution $q_0(\ve{x}_0)$. The forward process injects Guassian noise $\epsilon_t \sim \mathcal{N}(0, \mathbf{I})$ to the clean data as follows:
\begin{equation}
    \begin{split}
    &\ve{x}_t = \alpha_t \ve{x}_0 + \sigma_t \epsilon_t,\\
    \end{split}
    \label{eq:dis}
\end{equation}
where $t \sim \mathcal{U}[\eta,1]$ ($\eta$ is a very small quantity defaulting to 1e-5) and $\alpha_t$ and $\sigma_t$ are components of a predefined noise schedule. $\sigma_t$ is monotonically increasing from $0$ to $1$ in all diffusion models, while $\alpha_t$ can either remain constant (\textit{e.g.}, EDM~\citep{karras2022elucidating}) or decrease monotonically (\textit{e.g.}, VP-SDE~\citep{sde} and Rectified Flow~\citep{iclr22_rect}) from $0$ to $1$. The forward process in Eq.~\ref{eq:dis} can be rewritten as the following stochastic differential equation (SDE):
\begin{equation} 
\begin{split} d\ve{x}_t = f(t)\ve{x}_tdt + g(t)\overline{\bm{\omega}}_t, 
\end{split} 
\label{eq:forward_process}
\end{equation}
where $f(t)$ and $g(t)$ denote the \textit{drift coefficient} of $\ve{x}_t$ and the \textit{diffusion coefficient} of $\ve{x}_t$, respectively. $\overline{\bm{\omega}}_t$ refers to a standard Wiener process. Eq.~\ref{eq:forward_process} has a corresponding ODE-based reverse process defined as: 
\begin{equation} 
\begin{split} 
d\ve{x}_t = f(t)\ve{x}_t - \frac{1}{2}g^2(t)\nabla_\ve{x}\log q_t(\ve{x}), 
\end{split}
\label{eq:rrrrr}
\end{equation}
where $\nabla_\ve{x} \log q_t(\ve{x})$ denotes the score function $\nabla_{\ve{x}_t} \log p_t(\ve{x}_t)$. Since $\nabla_\ve{x} \log q_t(\ve{x})$ cannot be accessed during the reverse process, it must be replaced by a linear transformation $\frac{\model(\ve{x}_t,t)}{-\sigma_t}$ using the noise estimation model $\model(\cdot,\cdot)$ or the score function estimation model $\ve{s}_\theta(\cdot,\cdot)$.

\paragraph{Video Diffusion Model vs. Image Diffusion Model.} The most critical differences between IDM and VDM lie in their training sets and model architectures. Regarding training sets, IDM typically utilizes large-scale, high-quality datasets, enabled by the increasing availability of both real and synthetic image datasets~\citep{schuhmann2022laion}. Conversely, the performance of VDMs is significantly constrained by the limited number of publicly available video datasets, such as Webvid-10M~\citep{bain2021frozen} and Pandas-70M~\citep{chen2024panda}, which contain low-quality real data. In terms of model architectures, several VDMs differ from their IDM counterparts because video data $\ve{x}_0 \in \mathbb{R}^{b \times \mathrm{c} \times \mathrm{t} \times \mathrm{h} \times \mathrm{w}}$ includes an additional time dimension $\mathrm{t}$, unlike image data $\ve{x}_0 \in \mathbb{R}^{b \times \mathrm{c} \times \mathrm{h} \times \mathrm{w}}$, where $\mathrm{b}$, $\mathrm{c}$, $\mathrm{h}$, and $\mathrm{w}$ refer to batch size, number of channels, height, and width, respectively. Therefore, most VDMs incorporate both spatial and temporal blocks, where the spatial block aligns with the module used in the mainstream IDMs, while the temporal block employs 3D convolutions~\citep{animatediff} or tokens in the time dimension~\citep{latte} to capture temporal information.

\paragraph{SDEdit.} SDEdit~\citep{meng2021sdedit} is an image editing method that produces a latent variable capable of reconstructing the input $\ve{x}_0$. It first injects Gaussian noise into $\ve{x}_0$, then performs editing in the latent space, and finally obtains pure samples through a standard reverse process. This approach provides a scheme for editing the latent variable using IDM at each step of sampling. Since SDEdit is nontrivial to invert, we employ DDIM-Inversion~\citep{ddiminversion} to map the latent $\ve{x}_t$ back to its previous counterpart $\ve{x}_{t+\Delta t}$, where $ 1-t \geq \Delta t > 0$.

\paragraph{DDIM \& DDIM-Inversion.} DDIM~\citep{ddim} serves as an efficient ODE-based sampler that iteratively refines the initial Gaussian noise $\ve{x}_T$, ultimately generating a ``clean'' video $\ve{x}_0$ that conforms to the data distribution $p_0(\ve{x}_0)$. A notable characteristic of DDIM is its adherence to \textit{deterministic sampling}. The synthesized data produced by DDIM can be expressed as follows:
\begin{equation} \begin{split} & \ve{x}_t = \pounds(\ve{x}_s) = \alpha_t \left(\frac{\ve{x}_s-\sigma_s\model(\ve{x}_s,s)}{\alpha_s}\right) +\sigma_t\model(\ve{x}_s, s). \ \end{split} \label{eq:ddim_f_i} \end{equation}
In this equation, $s$ and $t$ denote timesteps $\in \mathcal{U}[0,1]$, with the condition that $t \leq s$. When under the constraint $t \geq s$, the standard DDIM in Eq.~\ref{eq:ddim_f_i} is referred to as DDIM-Inversion $\ve{x}_t = \pounds^{-\!1}(\ve{x}_s)$. In practical, DDIM is often incorporated into classifier-free guidance for more precise control as well as higher quality data synthesis. Thus, Eq.~\ref{eq:ddim_f_i} can be transferred into
\begin{equation}
\fontsize{9pt}{12pt}\selectfont
    \begin{split}
        &\ve{x}_t = \pounds(\ve{x}_s,\omega) = \Big[\alpha_t\ve{x}_s+(\alpha_s\sigma_t-\alpha_t\sigma_s)\\
        &[(\omega+1)\model(\ve{x}_s,s,\ve{c})-\omega\model(\ve{x}_s,s,\varnothing)]\Big]/\alpha_s, \\
    \end{split}
    \label{eq:cfg_ddim}
\end{equation}
where $\ve{c}$, $\varnothing$ and $\omega$ stand for the text prompt, the null prompt and the CFG scale, respectively.

\section{Approach}

Observing that IDMs produce high-quality samples without ensuring temporal coherence, while VDMs ensure temporal continuity but generate low-quality video, we propose \IVS\ to combine the strengths of both IDMs and VDMs (see Fig.~\ref{figure:illustration}). In this section, we first describe how to sample from $\ve{x}_t$ to $\ve{x}_{t+\Delta t}$ and from $\ve{x}_{t+\Delta t}$ to $\ve{x}_{t}$ using IDM and VDM. We then introduce \IVS\ and outline its design space of hyperparameters, followed by a theoretical analysis. Finally, we discuss the effect of sampling in the latent space on \IVS.

\subsection{Forward (Go!!) and Reverse (Back!!)}

A crucial step in overcoming the bottleneck of VDM $\VDM$ with IDM $\IDM$ is to address the domain gap between IDM and VDM caused by differences in training data. Therefore, a specialized rescheduling paradigm is required to achieve the mapping $\ve{x}_t \rightarrow \pounds^{-\!1}(\ve{x}_t) \rightarrow \pounds(\pounds^{-\!1}(\ve{x}_t)) = \ve{x}_t$, where $\pounds(\cdot)$ is a \textit{deterministic} function with an inverse $\pounds^{-\!1}(\cdot)$. Given this, we can modify $\ve{x}_{x+\Delta t}\approx \pounds^{-\!1}(\ve{x}_t)$ without disrupting the standard sampling process. We utilize DDIM-Inversion and DDIM to implement $\pounds^{-\!1}(\cdot,\cdot)$ and $\pounds(\cdot,\cdot)$, respectively. By introducing CFG, we can define a new paradigm to implement IDMs' information injection via the operator $\mathtt{G}$, \textit{i.e.} semantic information injection:
\begin{equation}
    \begin{split}
       \ve{x}^\prime_t & =  \mathcal{I}(\ve{x}_t,\omega_\textrm{go},\omega_\textrm{back}, \mathtt{G},\epsilon^\textrm{go}_\theta,\epsilon^\textrm{back}_\theta) =  \pounds(\ve{u}+\mathtt{G}(\ve{u}),\omega_\textrm{back}), \\
       &\text{ where }\ve{u} =  \pounds^{-\!1}(\ve{x}_t,\omega_\textrm{go}).\\
    \end{split}
    \label{eq:go_and_back_1}
\end{equation}\input{tables/chronomagic_150}
\input{tables/chronomagic_1649}In Eq.~\ref{eq:go_and_back_1}, $\omega_\textrm{go}$ and $\omega_\textrm{back}$ stand for the CFG scales for the DDIM-Inversion and DDIM, respectively. $\epsilon^\textrm{go}_\theta(\cdot,\cdot)$ and $\epsilon^\textrm{back}_\theta(\cdot,\cdot)$ are the noise estimation models used to perform $\pounds^{-\!1}(\cdot,\cdot)$ and $\pounds(\cdot,\cdot)$, respectively. $\mathtt{G}$ is any function whose mission is to make modifications to $\ve{u}$. It is worth noting that for IDMs, we first reshape $\ve{x}_t$ from the shape $b\!\times\!c\!\times\!t\!\times\!h\!\times\!w$ to $(b\!\times\!t)\!\times\!c\!\times\!h\!\times\!w$, and then pass it through the noise estimation model to perform the operation $\mathcal{I}$ in the practical implementation. If $\omega_\textrm{go}\equiv\omega_\textrm{back}$, $\epsilon^\textrm{go}_\theta(\cdot,\cdot)\equiv\epsilon^\textrm{back}_\theta(\cdot,\cdot)$ and $\mathtt{G}(\cdot)$ is an identity operator, then $\ve{x}^\prime_t\equiv\ve{x}_t$. To understand how $\mathcal{I}$ injects semantic information, we rewrite Eq.~\ref{eq:go_and_back_2} using a first-order Taylor expansion as\begin{equation}
\fontsize{6pt}{11pt}\selectfont
    \begin{split}
        &\mathcal{I}(\ve{x}_t,\omega_\textrm{go},\omega_\textrm{back},\mathtt{G},\epsilon^\textrm{go}_\theta,\epsilon^\textrm{back}_\theta) = \pounds(\pounds^{-\!1}(\ve{x}_t,\omega_\textrm{go})+\mathtt{G}(\pounds^{-\!1}(\ve{x}_t,\omega_\textrm{go})),\omega_\textrm{back}) \\
        &= \pounds(\pounds^{-\!1}(\ve{x}_t,\omega_\textrm{go}),\omega_\textrm{go}) + (\omega_\textrm{back}-\omega_\textrm{go})\frac{\partial\pounds(\pounds^{-\!1}(\ve{x}_t,\omega_\textrm{go}),\omega_\textrm{go})}{\partial \omega_\textrm{go}} \\
        &+ \mathtt{G}(\pounds^{-\!1}(\ve{x}_t,\omega_\textrm{go}))\frac{\partial\pounds(\pounds^{-\!1}(\ve{x}_t,\omega_\textrm{go}),\omega_\textrm{go})}{\partial \pounds^{-\!1}(\ve{x}_t,\omega_\textrm{go})} + \mathcal{O}((\omega_\textrm{back}-\omega_\textrm{go})^2) \\
        & + \mathcal{O}(\mathtt{G}(\pounds^{-\!1}(\ve{x}_t,\omega_\textrm{go}))^2) = \ve{x}_t \text{\textcolor{C2}{\quad\# define }}\textcolor{C2}{J=\alpha_{t+\Delta t}\sigma_t-\alpha_t\sigma_{t+\Delta t}}\\
        & + \textcolor{C4}{(\omega_\textrm{back}-\omega_\textrm{go})\Big[\partial \Big(\ve{x}_t + [J[(\omega_\textrm{back}+1)\epsilon^\textrm{back}_\theta(\ve{x}_{t+\Delta t},t + \Delta t,\ve{c})} \\
        &\textcolor{C4}{-\omega_\textrm{back}\epsilon^\textrm{back}_\theta(\ve{x}_{t+\Delta t},t + \Delta t,\varnothing)]/\alpha_{t+\Delta t}]\Big)\Big]/\partial \omega_\textrm{go}} \\
        & + (\omega_\textrm{back}-\omega_\textrm{go})\Big[\partial \Big((\alpha_{t}\sigma_{t+\Delta t}-\alpha_{t + \Delta t}\sigma_{t})[(\omega_\textrm{go}+1)\epsilon^\textrm{go}_\theta(\ve{x}_t,t,\ve{c}) \\
        &-\omega_\textrm{go}\epsilon^\textrm{go}_\theta(\ve{x}_t,t,\varnothing)]/\alpha_{t+\Delta t}\Big)\Big]/\partial \omega_\textrm{go}\\
        &  +\mathtt{G}(\pounds^{-\!1}(\ve{x}_t,\omega_\textrm{go}))\frac{\partial\pounds(\pounds^{-\!1}(\ve{x}_t,\omega_\textrm{go}),\omega_\textrm{go})}{\partial \pounds^{-\!1}(\ve{x}_t,\omega_\textrm{go})} \\
        &= \ve{x}_t + \textcolor{C6}{(\omega_\textrm{back}-\omega_\textrm{go})\frac{(\alpha_{t}\sigma_{t+\Delta t}-\alpha_{t+\Delta t}\sigma_{t})[\epsilon^\textrm{go}_\theta(\ve{x}_t,t,\ve{c})-\epsilon^\textrm{go}_\theta(\ve{x}_t,t,\varnothing)]}{\alpha_{t+\Delta t}}} \\
        &+ \mathtt{G}(\pounds^{-\!1}(\ve{x}_t,\omega_\textrm{go}))\frac{\partial\pounds(\pounds^{-\!1}(\ve{x}_t,\omega_\textrm{go}),\omega_\textrm{go})}{\partial \pounds^{-\!1}(\ve{x}_t,\omega_\textrm{go})}. \text{\textcolor{C4}{\quad\# Ignore second-order and higher terms.}}
    \end{split}
    \label{eq:go_and_back_2}
\end{equation}\begin{tikzpicture}[remember picture, overlay]
    \draw[thick, sloped, C4] (0.1,4.4) -- (6.8,5.3);
\end{tikzpicture}Observing the \textcolor{C6}{pink term} in Eq.~\ref{eq:go_and_back_2}, we can inject semantic information from both $\epsilon^\textrm{go}_\theta(\cdot,\cdot)$ and $\epsilon^\textrm{back}_\theta(\cdot,\cdot)$ into $\ve{x}_t$ by setting $\omega_\textrm{back}-\omega_\textrm{go}>0$. In particular, we set $\omega_\textrm{back}=-\omega_\textrm{go}$ and $\omega_\textrm{back}>0$ by default. Given this, we only need to replace $\epsilon^\textrm{go}_\theta$ with $\IDM$ or $\VDM$ to inject specific semantic information, thereby enhancing the visual quality or temporal coherence of the synthesized video.

\subsection{IV-mixed Sampler}
\label{sec:iv_mixed_sampler_sub}

\input{tables/traditional_bench_ucf}
\input{tables/traditional_bench_msr}

The paradigm defined in Eq.~\ref{eq:go_and_back_1} enables us to easily describe all the algorithms shown in Fig.~\ref{figure:motivation} that do not use Gaussian noise for score function estimation. For instance, ``I-I'' and ``V-V'' can represent $\mathcal{I}(\ve{x}_t,-h,h, \mathtt{G},\model^\mathrm{I},\model^\mathrm{I})$ and $\mathcal{I}(\ve{x}_t,-h,h, \mathtt{G},\model^\mathrm{V},\model^\mathrm{V})$, respectively, under the condition $h\!>\!0$ and $\mathtt{G}$ is an identity operator. Clearly, this represents only a single-step injection of semantic information. We can construct \IVS\ to allow multi-step injections of semantic information using $\mathtt{G}$ and the recursive definition:\begin{equation}
\setlength{\jot}{1pt}
    \begin{split}
        \ve{x}^\prime_t =\ &\mathtt{G}^1(\ve{x}_t) = \mathcal{I}(\ve{x}_t,-h,h, \mathtt{G}^2,\model^{1,\textrm{go}}, \model^{1,\textrm{back}}), \\
        & \mathtt{G}^2(\ve{y}) = \mathcal{I}(\ve{y},-h,h, \mathtt{G}^3,\model^{2,\textrm{go}}, \model^{2,\textrm{back}}), \\
        & \mathbf{\small \vdots} \\
        & \mathtt{G}^{N}(\ve{y}) = \mathcal{I}(\ve{y},-h,h, \mathtt{G}^{N+1},\model^{N,\textrm{go}}, \model^{N,\textrm{back}}),\\
        &\text{\quad s.t.\quad} N\geq 1, \\
    \end{split}
    \label{eq:general_definition}
\end{equation}
where $N$ and $\mathtt{G}^{N+1}$ refer to the number of semantic information injection and the identity operator, respectively. Through Eq.~\ref{eq:general_definition}, we can easily represent \IVS\ with different $N$. For instance, following the definition in Fig.~\ref{figure:motivation}, ``IV-VI'' can be described as $\ve{x}^\prime_t = \mathtt{G}^1(\ve{x}_t)= \mathcal{I}(\ve{x}_t,-h,h,\mathtt{G}^2,\model^{\mathrm{I}},\model^{\mathrm{I}})$, $\mathtt{G}^2(\ve{y})= \mathcal{I}(\ve{y},-h,h,\mathtt{G}^3,\model^{\mathrm{V}},\model^{\mathrm{V}})$ and $\mathtt{G}^3$ is an identity operator. Considering the computational overhead and performance trade-offs, this paper focuses only on the scenario where $N=2$.  In Sec.~\ref{sec:experiment}'s ablation study, we further restrict $\model^{1,\textrm{go}}$, $\model^{1,\textrm{back}}$, $\model^{2,\textrm{go}}$, and $\model^{2,\textrm{back}}$ to two models each, occupied by $\IDM$ and $\VDM$ (\textit{w.r.t.}, $C_2^4$ combinations), and find that ``IV-IV'' performs best. The visualization in Fig.~\ref{figure:visualization_main_paper} demonstrates that ``IV-IV'' significantly improves both the visual quality of the synthesized video and the consistency between the video and the text prompt.
\setlength{\jot}{\originaljot}

\subsection{Discussion}
\label{sec:discussion}

\paragraph{Hyperparameter Design Space.} In this paper, we elucidate three design choices: the $C_2^4$ combinations mentioned in Sec.~\ref{sec:iv_mixed_sampler_sub}, the intervals at which \IVS\ is performed during standard DDIM sampling, and the dynamic CFG scale. All three forms are expected to find the empirically optimal solution, with the first two being natural explorations and the last addressing the domain gap between IDMs and VDMs, which is expected to be mitigated by adjusting the CFG scale. For the last one, to be specific, we re-express $-h$ and $h$ in Eq.~\ref{eq:general_definition} as $-h^\textrm{go}(t)$ and $h^\textrm{back}(t)$. Inspired by the Karras's noise schedule~\citep{karras2022elucidating}, we define $h^\textrm{go}(t)$ and $h^\textrm{back}(t)$ as\begin{equation}
\fontsize{8pt}{11pt}\selectfont
    \begin{split}
        & h^\textrm{go}(t) = ({\left[\gamma^\textrm{begin}_\textrm{go}\right]}^{\frac{1}{\rho}}+t({\left[\gamma^\textrm{end}_\textrm{go}\right]}^{\frac{1}{\rho}}-{\left[\gamma^\textrm{begin}_\textrm{go}\right]}^{\frac{1}{\rho}}))^\rho, \\
        & h^\textrm{back}(t) = ({\left[\gamma^\textrm{begin}_\textrm{back}\right]}^{\frac{1}{\rho}}+t({\left[\gamma^\textrm{end}_\textrm{back}\right]}^{\frac{1}{\rho}}-{\left[\gamma^\textrm{begin}_\textrm{back}\right]}^{\frac{1}{\rho}}))^\rho, \\
    \end{split}
    \label{eq:dynamic_cfg_scale}
\end{equation}
where $\gamma^\textrm{end}_\textrm{go}$ and $\gamma^\textrm{end}_\textrm{back}$ represent the CFG scales when $t=1$, while $\gamma^\textrm{begin}_\textrm{go}$ and $\gamma^\textrm{begin}_\textrm{back}$ denote the CFG scales when $t=0$. The parameter $\rho$ controls the concave and convex properties of the CFG scale curve with respect to $t$. The experiments in Sec.~\ref{sec:experiment} demonstrate that the dynamic CFG scale can be adjusted to achieve performance improvements for specific aspects (\textit{e.g.}, semantic faithfulness).

\paragraph{Theoretical Analysis.} As shown in Theorem~\ref{the:111}, \IVS\ can be elegantly transformed into an ODE, taking the same form as Eq.~\ref{eq:rrrrr}. Consequently, \IVS\ preserves the standard sampling process (\textit{e.g.}, DDIM or Euler–Maruyama), enabling a trade-off between temporal coherence and visual quality by adjusting the parameters $\omega^\textrm{IDM}_\textrm{go-back}$, $\omega^\textrm{VDM}_\textrm{go-back}$, and $\omega$.

\begin{theorem}
\label{the:111}
(the proof in Appendix~\ref{apd:theoretical}) \IVS\ can be transferred to an ODE. For example, the ODE corresponding to ``IV-IV'' is
\begin{equation}
\begin{split}
&d\ve{x}_t = f(t)\ve{x}_t - \frac{1}{2}g^2(t)\Big[\omega^\textrm{IDM}_\textrm{go-back}\frac{g^2(t)}{2}\nabla_\ve{x}\log q^\textrm{IDM}_t(\ve{c}|\ve{x})\\
&+(\omega^\textrm{VDM}_\textrm{go-back}+\omega)\frac{g^2(t)}{2}\nabla_\ve{x}\log q^\textrm{VDM}_{t}(\ve{c}|\ve{x})\Big],\\
\end{split}
\end{equation}
Here, $\omega$ refers to the vanilla CFG scale, while both $\omega^\textrm{IDM}_\textrm{go-back}$ and $\omega^\textrm{IDM}_\textrm{go-back}$ are CFG scales that are greater than 0. Let $\nabla_\ve{x}\log q^\textrm{IDM}_{t}(\ve{c}|\ve{x})$ and $\nabla_\ve{x}\log q^\textrm{VDM}_{t}(\ve{c}|\ve{x})$ represent the score function estimated for $p_t(\ve{x}_t)$ using IDM and VDM under classifier-free guidance.
\end{theorem}

\begin{figure*}[!t]
\centering
\includegraphics[width=1.0\textwidth,trim={0cm 0cm 0cm 0cm},clip]{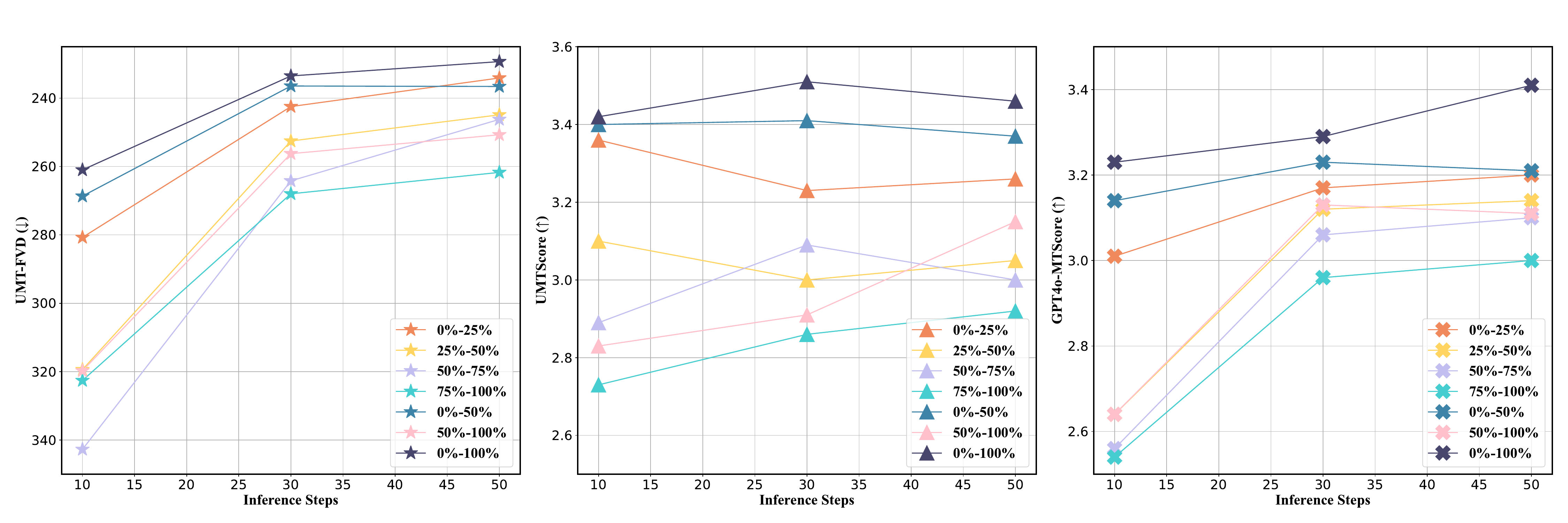}
\vspace{-15pt}
\caption{\textbf{Ablation studies on sampling intervals of \IVS\ (``IV-IV'') with Animatediff (SD V1.5, Motion Adapter V3).} The \textcolor{C3}{\textsc{Begin}}\%-\textcolor{C3}{\textsc{End}}\% in the legend indicates the portion of the entire sampling process performed by \IVS. For example, in a 50-step sampling scenario, 0\%-50\% corresponds to \IVS\ being applied during steps 1-25. More details of ``IV-VI'' can be found in Appendix~\ref{apd:additional_ablation_study}.}
\vspace{-5pt}
\label{figure:ablation_study_sampling_interval}
\end{figure*}

\begin{figure*}[!t]
\vspace{0pt}
\centering
\includegraphics[width=1.0\textwidth,trim={0cm 0cm 0cm 0cm},clip]{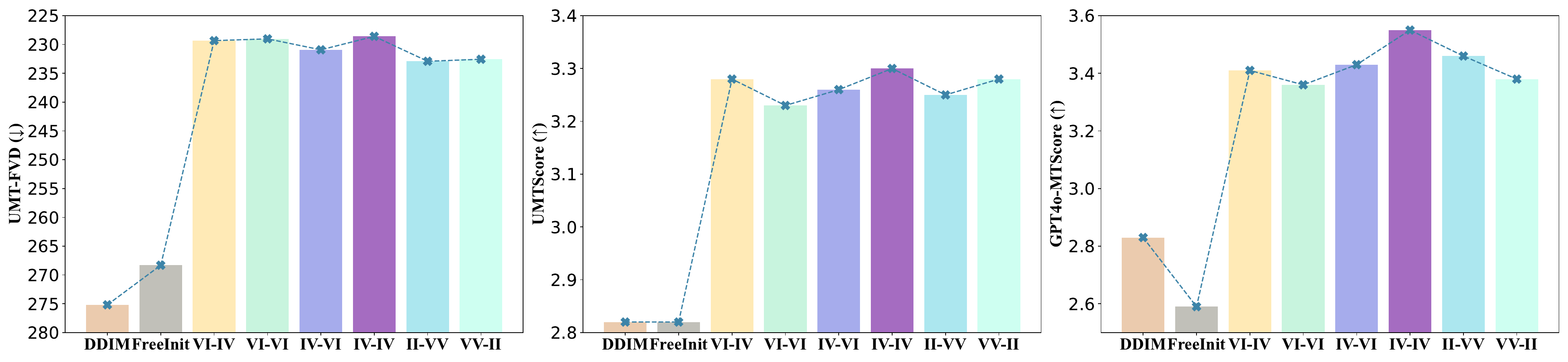}
\vspace{-15pt}
\caption{\textbf{Ablation studies on different $C_4^2$ species combinations with Animatediff (SD V1.5, Motion Adapter V3).} We can clearly observe that \IVS\ (``IV-IV'') is the winner across all metrics.}
\vspace{-15pt}
\label{figure:c42}
\end{figure*}

\begin{figure*}[t]
\centering
\includegraphics[width=1.0\textwidth,trim={0cm 0cm 0cm 0cm},clip]{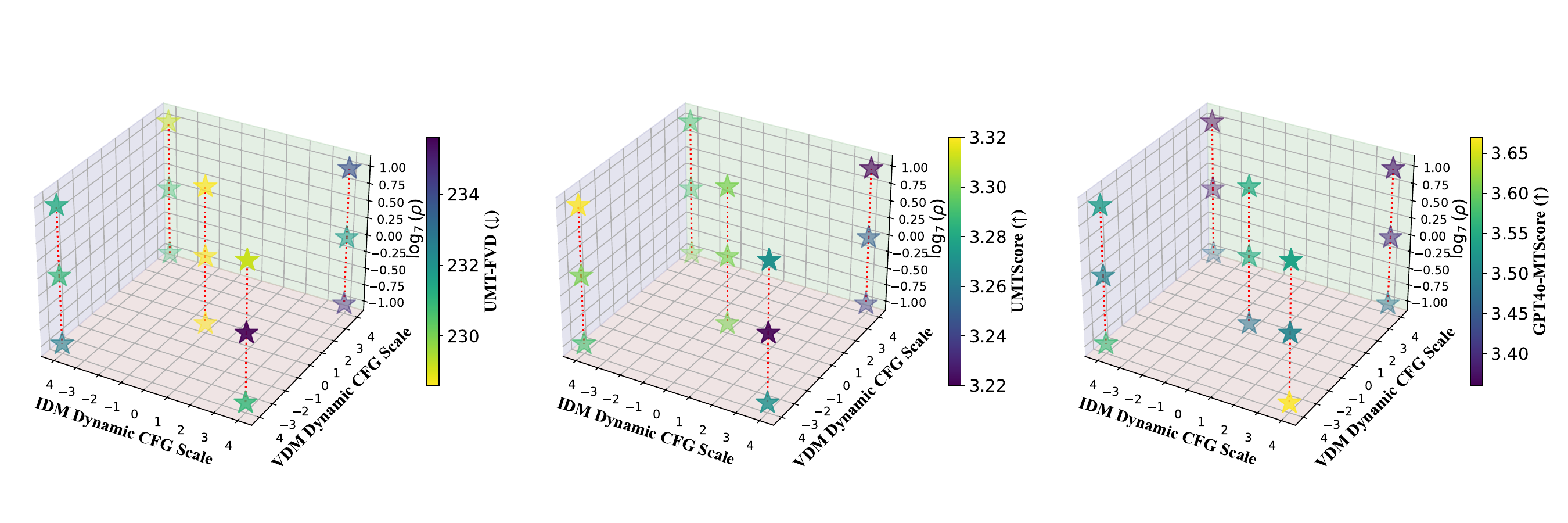}
\vspace{-15pt}
\caption{\textbf{Ablation studies on dynamic CFG scale with Animatediff (SD V1.5, Motion Adapter V3).} We ensure that $h^\textrm{go}(t) \equiv h^\textrm{back}(t)$. Consequently, the IDM dynamic CFG scale corresponds to IDM's $\gamma_\textrm{go}^\textrm{end}-\gamma_\textrm{go}^\textrm{begin}$. Similarly, the VDM dynamic CFG scale corresponds to VDM's $\gamma_\textrm{go}^\textrm{end}-\gamma_\textrm{go}^\textrm{begin}$. $\log_7(\rho)$ is 1, 0, and -1 indicates whether the change from $\gamma_\textrm{go}^\textrm{end}$ to $\gamma_\textrm{go}^\textrm{begin}$ is convex, linear, or concave, respectively.}
\vspace{-15pt}
\label{figure:dynamic_cfg_scale}
\end{figure*}

\paragraph{The Influence of Latent Space.} Most current high-resolution IDMs and VDMs follow the latent diffusion model (LDM) paradigm~\citep{SDV1.5}. As a result, leveraging \IVS\ requires IDM and VDM to share the same latent space, meaning they must use the same VAE~\citep{vae_1}. Fortunately, for most VDMs, we always can find the corresponding IDM that share the same latent space with the VDM. For cases where this condition is not met, we can convert the video from shape b$\times$c$\times$t$\times$h$\times$w to (b$\times$t)$\times$c$\times$1$\times$h$\times$w and then pass it through VDM, which is referred to as using ``IDM''. This paradigm is also applied in I4VGen~\citep{guo2024i4vgen}.

\section{Experiment}
\label{sec:experiment}
In this section, we present experiments to demonstrate the effectiveness of \IVS. Specifically, we perform qualitative and quantitative comparisons across various benchmarks on multiple T2V diffusion models, including ModelScope-T2V~\citep{modelscope}, Animatediff~\citep{animatediff}, and VideoCrafterV2~\citep{chen2023videocrafter1}. Additionally, we compare \IVS\ with three heavy-inference algorithms, FreeInit~\citep{freeinit}, Unictrl~\citep{chen2024unictrl} and I4VGen~\citep{guo2024i4vgen}, on VDM. Note that I4VGen is also an algorithm designed to enhance VDM performance using IDM. However, its IDM is configured to be consistent with VDM, achieving image synthesis by reshaping from b$\times$c$\times$t$\times$h$\times$w to (b$\times$t)$\times$c$\times$1$\times$h$\times$w before passing it through VDM. For further details, please refer to Appendix~\ref{sec:heavy_inference}. Finally, we conduct extensive ablation studies and present visualization to validate the optimal solution of various design choices. More implementation details can be found in Appendix~\ref{apd:implementation_details}.

\subsection{Main Results}
\label{sec:main_results} 

\paragraph{Chronomagic-Bench 150 \& 1649.} We evaluate the effectiveness of \IVS\ on three different VDMs: VideoCrafterV2, ModelScope-T2V, and Animatediff (SD V1.5, Motion Adapter V3). The comparison results on Chronomagic-Bench-150 (\textit{w.r.t}, 150 prompts) are presented in Table~\ref{tab:comparison}. We employ three metrics in this benchmark: UMT-FVD (for visual quality), UMTScore (for semantic faithfulness), and GPT4o-MTScore (for temporal coherence and metamorphic amplitude) to assess our proposed method. The experimental results in Table~\ref{tab:comparison} show that \IVS\ significantly outperforms both the standard sampling method and other computationally intensive algorithms. This indicates that rational integration of IDM and VDM has great potential to improve the performance of video synthesis tasks. Although \IVS\ does not outperform all comparative methods on Animatediff and ModelScope-T2V in terms of UMT-FVD, it still shows a significant improvement over standard sampling algorithms. It is worth noting that ModelScope-T2V was unable to locate a high-quality IDM due to its low resolution (\textit{i.e.}, 224$\times$224) synthesized video. In contrast, \IVS\ do not perform best on Animatediff because I4VGen is an algorithm that integrates both IDM and VDM. Moreover, on the more comprehensive Chronomagic-Bench-1649 (\textit{w.r.t}, 1649 prompts), \IVS\ outperforms all comparative methods across all metrics and models. As illustrated in Table~\ref{tab:comparison_1649}, \IVS\ achieves the best performance on all metrics, even when compared to the latest SOTA method, I4VGen. These experimental results underscore the strong generalization capabilities of \IVS, highlighting its effectiveness as a plug-and-play solution that can be seamlessly integrated across various VDMs.

\paragraph{UCF-101-FVD \& MSR-VTT-FVD.} Additionally, to complement our experimental results, we conduct experiments on several traditional benchmarks, including UCF-101-FVD and MSR-VTT-FVD. The FVD results for UCF-101 and MSR-VTT are presented in Table~\ref{tab:traditional_bench_animatediff} and Table~\ref{tab:traditional_bench_modelscope}, respectively. These quantitative results significantly substantiate the superiority of \IVS, highlighting its effectiveness in outperforming existing heavy-inference approaches on VDMs. Through these results and visualized in Fig.~\ref{figure:visualization_main_paper}, \IVS\ has strong generalization ability across different VDMs, which possesses significant practical application value in real-world scenarios.

\subsection{Ablation Studies} As described in Sec.~\ref{sec:discussion}, we elucidate three design choices, namely the sampling interval of \IVS, the $C_4^2$ species combinations, and the dynamic CFG scale. For \textcolor{C3}{\textit{1)} the sampling interval of \IVS}, the results in Fig.~\ref{figure:ablation_study_sampling_interval} clearly illustrate that performing \IVS\ across all sampling steps is optimal. Another evident conclusion is that the closer the sampling interval of \IVS\ is to $t\!=\!1$, the more significant the performance gain. This suggests that to save computational overhead, \IVS\ can be applied within the 0\%-50\% sampling interval or even restricted to the 0\%-25\% interval. For \textcolor{C3}{\textit{2)} the $C_4^2$ species combinations}, we present its ablation results in Fig.~\ref{figure:c42}. It is evident that ``IV-IV'' outperforms all metrics and significantly surpasses both FreeInit and standard DDIM. This suggests that there is an empirically optimal combination of results within \IVS. Accordingly, we use ``IV-IV'' for all comparison experiments. For \textcolor{C3}{\textit{3)} the dynamic CFG scale}, the conclusions are not as intuitive as the first two design choices. Specifically, we considered a total of 15 combinations of dynamic CFG scales, where the CFG scale varies from $\gamma_\textrm{go}^\textrm{end}$ to $\gamma_\textrm{go}^\textrm{begin}$ across five convex, five straight, and five concave functions. In Fig.~\ref{figure:dynamic_cfg_scale}, we set $\rho$ to 7, 1, and ${1/7}$\footnote{$\rho=7$ is the default implementation of Karras's noise schedule~\citep{karras2022elucidating}.} to model changes in the CFG scale as convex, straight, and concave functions, respectively. For each specified $\rho$, we consider five combinations of the IDM's CFG scale $\omega_\textrm{go-back}^\textrm{IDM}$ and VDM's CFG scale $\omega_\textrm{go-back}^\textrm{VDM}$: (1) both remaining constant, (2) $\omega_\textrm{go-back}^\textrm{IDM}$ increasing and $\omega_\textrm{go-back}^\textrm{VDM}$ increasing, (3) $\omega_\textrm{go-back}^\textrm{IDM}$ decreasing and $\omega_\textrm{go-back}^\textrm{VDM}$ decreasing, (4) $\omega_\textrm{go-back}^\textrm{IDM}$ increasing and $\omega_\textrm{go-back}^\textrm{VDM}$ decreasing, and (5) $\omega_\textrm{go-back}^\textrm{IDM}$ decreasing and $\omega_\textrm{go-back}^\textrm{VDM}$ increasing. For the ``constant'' case we make $\gamma_\textrm{go}^\textrm{end}=\gamma_\textrm{go}^\textrm{begin}=4$, for the ``decreasing'' case we make $\gamma_\textrm{go}^\textrm{end}=6,\ \gamma_\textrm{go}^\textrm{begin}=2$, and for the ``increasing'' we make $\gamma_\textrm{go}^\textrm{end}=2,\ \gamma_\textrm{go}^\textrm{begin}=6$. As illustrated in Fig.~\ref{figure:dynamic_cfg_scale}, we find that for visual quality (\textit{w.r.t.}, UMT-FVD), keeping both $\omega_\textrm{go-back}^\textrm{IDM}$ and $\omega_\textrm{go-back}^\textrm{VDM}$ constant is the best choice. For semantic faithfulness (\textit{w.r.t.}, UMTScore), the optimal strategy is to have both $\omega_\textrm{go-back}^\textrm{IDM}$ and $\omega_\textrm{go-back}^\textrm{VDM}$ increasing. For temporal coherence (\textit{w.r.t.}, GPT4o-MTScore), it is optimal for $\omega_\textrm{go-back}^\textrm{IDM}$ to decrease while $\omega_\textrm{go-back}^\textrm{VDM}$ increases. Therefore, to enhance the performance of a specific aspect of synthesized video, we can adjust the dynamic CFG scale to achieve the empirically optimal trade-off.

\paragraph{Visualization.} We visualize the standard sampling and \IVS\ of the synthesized video in Fig.~\ref{figure:visualization_main_paper}. It can be observed that \IVS\ significantly improves both visual quality and semantic faithfulness. In addition to this, we empirically invited a number of other AIGC-related researchers to judge the video quality and agreed that \IVS's enhancement could be observed by the naked eye.

\section{Limitation}

Althrough \IVS\ significantly improves the performance of VDM, it introduces additional computational costs. For "IV-IV" on Animatediff, it increases the number of function evaluation (NFE) from 50 to 250. In the practical implementation, the computational overhead went up from 21s to 92s at a single RTX 4090 GPU. This problem could potentially be addressed in the future by distillation algorithms similar to accelerated sampling~\citep{icml23_consistency,shao2023catch,iclr22_progressive}. This exploration of \textit{\textsc{Inference Scaling Laws}} first, and then distilling the performance gains it achieves back to the foundation model may be a viable path for the future.

\section{Conclusion}
In this paper, we propose \IVS\ to enhance the visual quality of synthesized videos by leveraging an IDM while ensuring temporal coherence through a VDM. The algorithm utilizes DDIM and DDIM-Inversion to correct latent representations $\ve{x}_t$ at any time point $t$, enabling seamless integration into any VDM and sampling interval. \IVS\ can be formulated as an ODE, achieving a trade-off between visual quality and temporal coherence by adjusting the CFG scales of both the IDM and VDM. In the future, we plan to fine-tune several stronger IDMs, such as FLUX, to better adapt the latent space of target VDMs, thereby further enhancing the performance of VDMs. We anticipate \IVS\ will be widely applicable in vision generation tasks.

\paragraph{Ethics Statement.} We present \IVS, a method designed to enhance the semantic accuracy and visual quality of video produced by Video Diffusion Models. Although our approach does not directly engage with real-world datasets, we are dedicated to ensuring the ethical use of prompts, while respecting user autonomy and striving for positive outcomes. Acknowledging the commercial potential of \IVS, we emphasize a responsible and ethical deployment of the technology, aiming to maximize societal benefits while carefully mitigating any potential risks.

\bibliography{main}
\bibliographystyle{ieeenat_fullname}

\clearpage
\setcounter{page}{1}
\onecolumn
\appendix

\section{Additional Implementation Details}
\label{apd:implementation_details}

\subsection{Benchmarks}
\label{apd:benchmark}
We present the relevant metrics and benchmarks used for comparison in the main paper..

\paragraph{UCF-101-related FVD.} The UCF-101 dataset is an action recognition dataset comprising 101 categories, with all videos sourced from Youtube. Each video has a fixed frame rate of 25 frames per second~(FPS) and a resolution of 320×240. Several previous works~\citep{blattmann2023stable,freeinit,chen2024unictrl} have validated the generation performance of VDMs on the UCF-101 dataset using Fréchet Video Distance~(FVD)~\citep{unterthiner2019fvd}. However, a comprehensive evaluation benchmark for UCF-101 is still lacking. To address this, we follow the methodology of {FreeInit}, utilizing the prompts listed in~\cite{ge2023preserve} to synthesize videos and assess inference performance with FVD. Specifically, we synthesize 5 videos for each of the 101 prompts provided by~\cite{ge2023preserve}, resulting in a total of 505 synthesized videos. We then compute the FVD between these 505 synthesized videos and 505 randomly sampled videos from the UCF-101 dataset~(5 per class), using the built-in FVD evaluation code from Open-Sora-Plan\footnote{\url{https://github.com/PKU-YuanGroup/Open-Sora-Plan}}.

\paragraph{MSR-VTT-related FVD.} The MSR-VTT dataset~\citep{msrvtt} is a large-scale dataset for open-domain video captioning, featuring 10,000 video clips categorized into 20 classes. The standard split of the MSR-VTT dataset includes 6,513 clips for training, 497 clips for validation, and 2,990 clips for testing. For our evaluation, we utilize all 497 validation videos. To ensure evaluation stability, we synthesize a total of 1,491 videos based on prompts from these validation videos, with each prompt producing 3 different videos. We assess the results using the built-in FVD evaluation code from Open-Sora-Plan.

\paragraph{Chronomagic-Bench-150.} Chronomagic-Bench-150, introduced in~\citep{yuan2024chronomagic} and recently accepted by NeurIPS 2024's dataset and benchmark track, serves as a comprehensive benchmark for metamorphic evaluation of timelapse T2V synthesis. This benchmark includes 4 main categories of time-lapse videos: biological, human-created, meteorological, and physical, further divided into 75 subcategories. Each subcategory contains two challenging prompts, leading to in a total of 150 prompts. We consider three distinct metrics in Chronomagic-Bench-150: {UMT-FVD (↓)}, {UMTScore (↑)}, and {GPT4o-MTScore (↑)}, each addressing different evaluation aspects. Specifically, {UMT-FVD (↓)}~\citep{UMT_FVD} leverages the UMT~\citep{li2023unmasked} feature space to compute FVD, assessing the visual quality of the synthesized video. {UMTScore (↑)} utilizes the UMT~\citep{li2023unmasked} feature space to compute CLIPScore~\citep{hessel2021clipscore}, evaluating the text relevance of the synthesized video. Lastly, {GPT4o-MTScore (↑)} is a fine-grained metric that employs GPT-4o~\citep{GPT4} as an evaluator, aligning with human perception to accurately reflect the metamorphic amplitude and temporal coherence of T2V models.

\paragraph{Chronomagic-Bench-1649.} Chronomagic-Bench-1649, introduced in~\citep{yuan2024chronomagic} and recently accepted by NeurIPS 2024's dataset and benchmark track, is a comprehensive benchmark designed for the metamorphic evaluation of timelapse T2V synthesis. While it shares 75 subcategories with Chronomagic-Bench-150, it offers a more extensive evaluation framework with 1649 prompts, making it significantly more comprehensive than its lightweight counterpart Chronomagic-Bench-150. Chronomagic-Bench-1649 includes 4 key metrics: {UMT-FVD (↓)}, {MTScore (↑)}, {UMTScore (↑)}, and {GPT4o-MTScore (↑)}, each serving to evaluate different aspects of video synthesis. Specifically, {UMT-FVD (↓)}~\citep{UMT_FVD} utilizes the UMT~\citep{li2023unmasked} feature space to compute FVD, assessing the visual quality of the synthesized videos. {MTScore (↑)} measures metamorphic amplitude, indicating the degree of change between frames. {UMTScore (↑)} leverages the UMT~\citep{li2023unmasked} feature space to compute CLIPScore~\citep{hessel2021clipscore}, evaluating the text relevance of the synthesized videos. Finally, {GPT4o-MTScore (↑)} is a fine-grained metric that employs GPT-4o~\citep{GPT4} as an evaluator, aligning with human perception to accurately reflect the metamorphic amplitude and temporal coherence of T2V models. As with to Chronomagic-Bench-150, we choose to ignore the \textrm{MTScore (↑)} metric in our experiments due to its limitations.

\subsection{Video Diffusion Models}
\label{apd:vdm}

We describe the VDMs utilized in this work. Specifically, we employ 3 VDMs with distinct architectures: {ModelScope-T2V}~\citep{modelscope}, {Animatediff}~\citep{animatediff}, {VideoCrafterV2}~\citep{chen2024videocrafter2overcomingdatalimitations}.

\paragraph{ModelScope-T2V.} {ModelScope-T2V} incorporates spatio-temporal blocks to ensure consistent frame generation and smooth motion transitions. Its key features include the utilization of 3D convolution andtraining from scratch. The input video size is structured as ${3\!\times\!16\!\times\!256\!\times\!256}$, where $3$ represents the number of channels, $16$ is the number of frames, and $256\!\times\!256$ indicates the resolution of each frame. This configuration allows the model to effectively capture both spatial adn temporal features, facilitating high-quality video synthesis.

\paragraph{Animatediff.} {Animatediff} does not require training from scratch; instead, it only needs fine-tuning on existing image diffusion models. Its motion adapter serves as a plug-and-play module, allowing most community text-to-image models to be transformed into animation generators. In this paper, we consider the latest version {Animatediff (SD V1.5, Motion Adapter V3)}, which is fine-tuned from {SD V1.5}. The input video size of {Animatediff (SD V1.5, Motion Adapter V3)} is ${3\!\times\!16\!\times\!512\!\times\!512}$, where $3$ indicates the number of channels, $16$ represents the number of frames, and $512\!\times\!512$ specifies the resolution. Note that there are differences in the performance of this VDM because we used a different resolution than the one used in the Chronomagic-Bench paper~\citep{yuan2024chronomagic}.

\paragraph{VideoCrafterV2.} {VideoCrafterV2} focuses on T2V synthesis, aiming to synthesize high-quality videos from prompts. This work investigates a training scheme for video models based on {Stable Diffusion}~\citep{sd}, exploring how to leverage low-quality videos and synthesized high-quality images to develop a superior video model.  The input video size of {VideoCrafterV2} is ${3\!\times\!16\!\times\!512\!\times\!320}$, where $3$ indicates the number of channels, $16$ represents the number of frames, and $512\!\times\!320$ specifies the resolution.

\subsection{Heavy-inference Algorithm on VDM}
\label{sec:heavy_inference}

Here we discuss two popular VDM-based heavy-inference algorithms FreeInit~\citep{freeinit} and {I4VGen}~\citep{guo2024i4vgen}.

\paragraph{I4VGen.} I4VGen~\citep{guo2024i4vgen} is a training-free and plug-and-play video diffusion inference framework that enhances text-to-video synthesis by leveraging robust image techniques. To be specific, I4VGen decomposes the process into two stages: anchor image synthesis and anchor image-guided video synthesis. A well-designed generation-selection pipeline is used to create visually realistic and semantically faithful anchor images, while score distillation sampling (SDS)~\citep{iclr2023_dreamfusion} is employed to animate the images into dynamic videos, followed by a video regeneration process to refine the output. In its official implementation, both phases are realized by VDM, where the anchor image synthesis is performed by merging the time dimension into the batch size dimension through VDM. In essence, I4VGen does not introduce true IDMs to improve the quality of synthesized video obtained from VDMs.

\paragraph{FreeInit.} FreeInit~\citep{freeinit} is a novel inference-time strategy designed to enhance temporal consistency in video generation using diffusion models. This approach addresses a key issue: the difference in the spatial-temporal frequency distribution of noise between training and inference, which leads to poor video quality. FreeInit iteratively refines the low-frequency components of the initial noise during inference, bridging this gap without requiring additional training.

\subsection{Hyperparameter Settings}
\label{apd:hyperparameter_setting}

\input{tables/appendix_chronomagic_150_modelscope}
\input{tables/chronomagic_150_videocrafterv2_appendix}
\begin{figure*}[!t]
\centering
\includegraphics[width=1.0\textwidth,trim={0cm 0cm 0cm 0cm},clip]{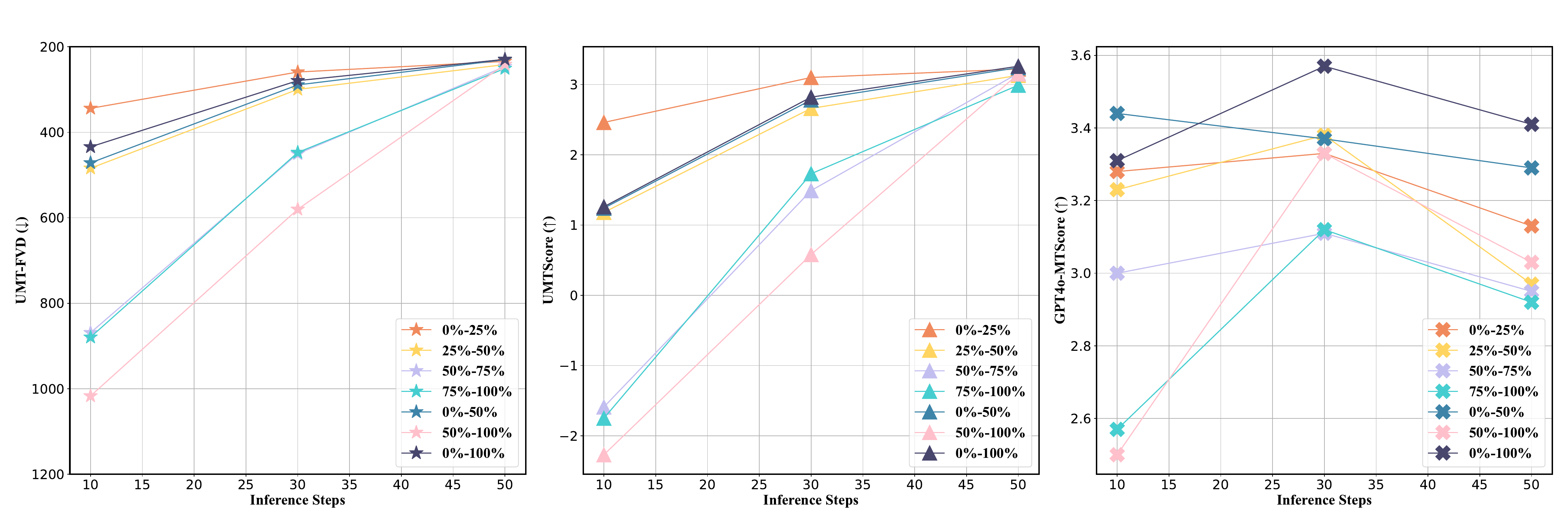}
\vspace{-12pt}
\caption{Ablation studies on sampling intervals of \IVS\ (``VI-IV''). The \textcolor{C2}{\textsc{Begin}}\%-\textcolor{C2}{\textsc{End}}\% in the legend indicates the portion of the entire sampling process performed by \IVS.}
\vspace{-15pt}
\label{figure:ablation_study_sampling_interval_ivvi}
\end{figure*}
\begin{figure*}[!t]
\centering
\includegraphics[width=1.0\textwidth,trim={0cm 0cm 0cm 0cm},clip]{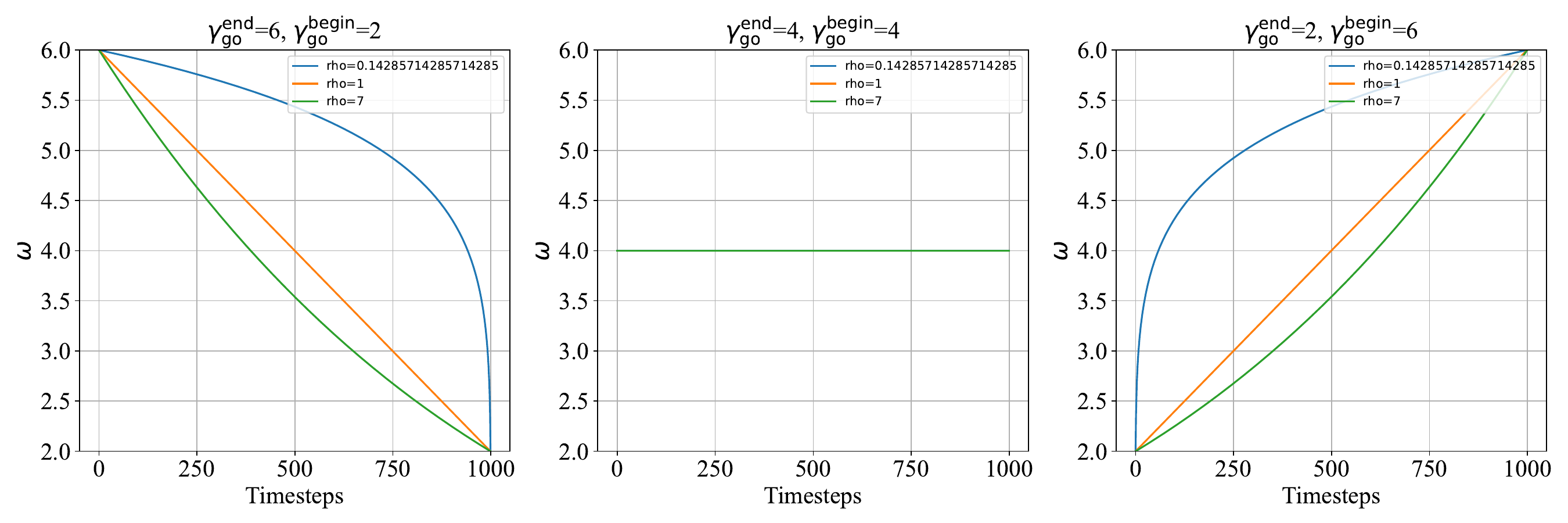}
\vspace{-12pt}
\caption{The visualization of Karras's noise schedule.}
\vspace{-15pt}
\label{figure:cfg_explanation}
\end{figure*}
For all comparison experiments, we used the form ``IV-IV'' and perform \IVS\ at all time steps of the standard DDIM sampling. In addition, $\gamma^\textrm{end}_\textrm{go}$, $\gamma^\textrm{end}_\textrm{back}$, $\gamma^\textrm{begin}_\textrm{go}$ and $\gamma^\textrm{begin}_\textrm{back}$ all are set as 4. For both Animatediff and ModelScope-T2V, we use stable diffusion (SD) V1.5 as the IDM. Note that we experimented with using Mini SD as the IDM for ModelScope-T2V to maintain a consistent resolution of 256×256. However, as illustrated in Table~\ref{tab:comparison_modelscope_appendix}, we found that its performance was inferior to using SD V1.5 with upsampling and downsampling. For VideoCrafterV2, we use Realistic Vision V6.0 B1~\citep{RealVis} as the IDM to accommodate a resolution of 512$\!\times\!$320. For the remaining configurations, we follow the sampling form recommended by the corresponding VDMs. Furthermore, we find that applying ``IV-IV'' at every step on VideoCrafterV2 destroys temporal coherence. Therefore, we replace ``IV-IV'' with ``VV-VV'' for z\%. The results of the ablation experiments are shown in Table~\ref{tab:comparison_z_percent}. We finally chose z\%=66.7\% as the final solution.

\section{Python-style Pseudo Code}
\label{apd:pseudo_code}

We present the python-style pseudo-code in Fig.~\ref{algo:ctc} to make it easier to understand \IVS.

\input{algorithms/iviv_sampler}

\section{Theoretical Proof}
\label{apd:theoretical}

Here we give the proof of Theorem~\ref{the:111} in the main paper. We use ``IV-IV'' for an example, and the derivation of other forms of \IVS\ is similar to ``IV-IV'' and is not described additionally. First, \IVS\ (\textit{w.r.t.}, ``IV-IV'') can be rewritten as
\begin{equation}
\begin{split}
        & \ve{x}^\prime_t = \ve{x}_t + \omega_\textrm{go}^1\frac{g^2(t)}{2}\nabla_\ve{x}\log q^\textrm{IDM}_t(\ve{c}|\ve{x}) +  \omega_\textrm{go}^2\frac{g^2(t+\eta)}{2}\nabla_\ve{x}\log q^\textrm{VDM}_{t+\eta}(\ve{c}|\ve{x}) \\
        &- \omega_\textrm{back}^2\frac{g^2(t+\eta)}{2}\nabla_\ve{x}\log q^\textrm{IDM}_{t+\eta}(\ve{c}|\ve{x}) - \omega_\textrm{back}^1\frac{g^2(t)}{2}\nabla_\ve{x}\log q^\textrm{VDM}_{t}(\ve{c}|\ve{x}), \\
\end{split}
\label{eq:apd:forward_backward_c1}
\end{equation}
where $\eta$ represents the sampling step, which is usually extremely small. $\nabla_\ve{x}\log q^\textrm{IDM}_{t}(\ve{c}|\ve{x})$ and $\nabla_\ve{x}\log q^\textrm{VDM}_{t}(\ve{c}|\ve{x})$ represent the score function estimated for $p_t(\ve{x}_t)$ using IDM and VDM under classifier-free guidance. Since $\eta$ is very small, and assume that $\omega^2_\textrm{back}$ is larger than $\omega^1_\textrm{go}$ and $\omega^1_\textrm{back}$ is larger than $\omega^2_\textrm{go}$, Eq.~\ref{eq:apd:forward_backward_c1} can be rewritten as
\begin{equation}
\begin{split}
        & \ve{x}^\prime_t = \ve{x}_t- \omega^\textrm{IDM}_\textrm{go-back}\frac{g^2(t)}{2}\nabla_\ve{x}\log q^\textrm{IDM}_t(\ve{c}|\ve{x})- \omega^\textrm{VDM}_\textrm{go-back}\frac{g^2(t)}{2}\nabla_\ve{x}\log q^\textrm{VDM}_{t}(\ve{c}|\ve{x}), \\
\end{split}
\label{eq:apd:forward_backward_c2}
\end{equation}
where $\omega^\textrm{IDM}_\textrm{go-back}$ and $\omega^\textrm{VDM}_\textrm{go-back}$ are large than $0$. Given this, \IVS\ can be written as an ordinary differential equation:
\begin{equation} 
\begin{split} 
d\ve{x}_t = f(t)\ve{x}_t - \frac{1}{2}g^2(t)\left[\omega^\textrm{IDM}_\textrm{go-back}\frac{g^2(t)}{2}\nabla_\ve{x}\log q^\textrm{IDM}_t(\ve{c}|\ve{x})+(\omega^\textrm{VDM}_\textrm{go-back}+\omega)\frac{g^2(t)}{2}\nabla_\ve{x}\log q^\textrm{VDM}_{t}(\ve{c}|\ve{x}),\right], 
\end{split} 
\end{equation}
where $\omega$ refers to the vanilla CFG scale.

\section{Additional Ablation Study}
\label{apd:additional_ablation_study}

We present Fig.~\ref{figure:ablation_study_sampling_interval_ivvi} here as a supplement to Fig~\ref{figure:ablation_study_sampling_interval} (\textit{w.r.t.}, the sampling interval of \IVS) in the main paper. As illuatrated in Fig.~\ref{figure:ablation_study_sampling_interval} and Fig.~\ref{figure:ablation_study_sampling_interval_ivvi}, it can be noticed that ``IV-IV'' performs significantly better than ``VI-IV'' under almost all settings. Furthermore, we visualize Karras's noise schedule of our proposed dynamic CFG scale in Fig.~\ref{figure:cfg_explanation} for clear  understanding.

\section{Visualization}
\begin{figure}[h]
\centering
\includegraphics[width=1.\textwidth,trim={0cm 0cm 0cm 0cm},clip]{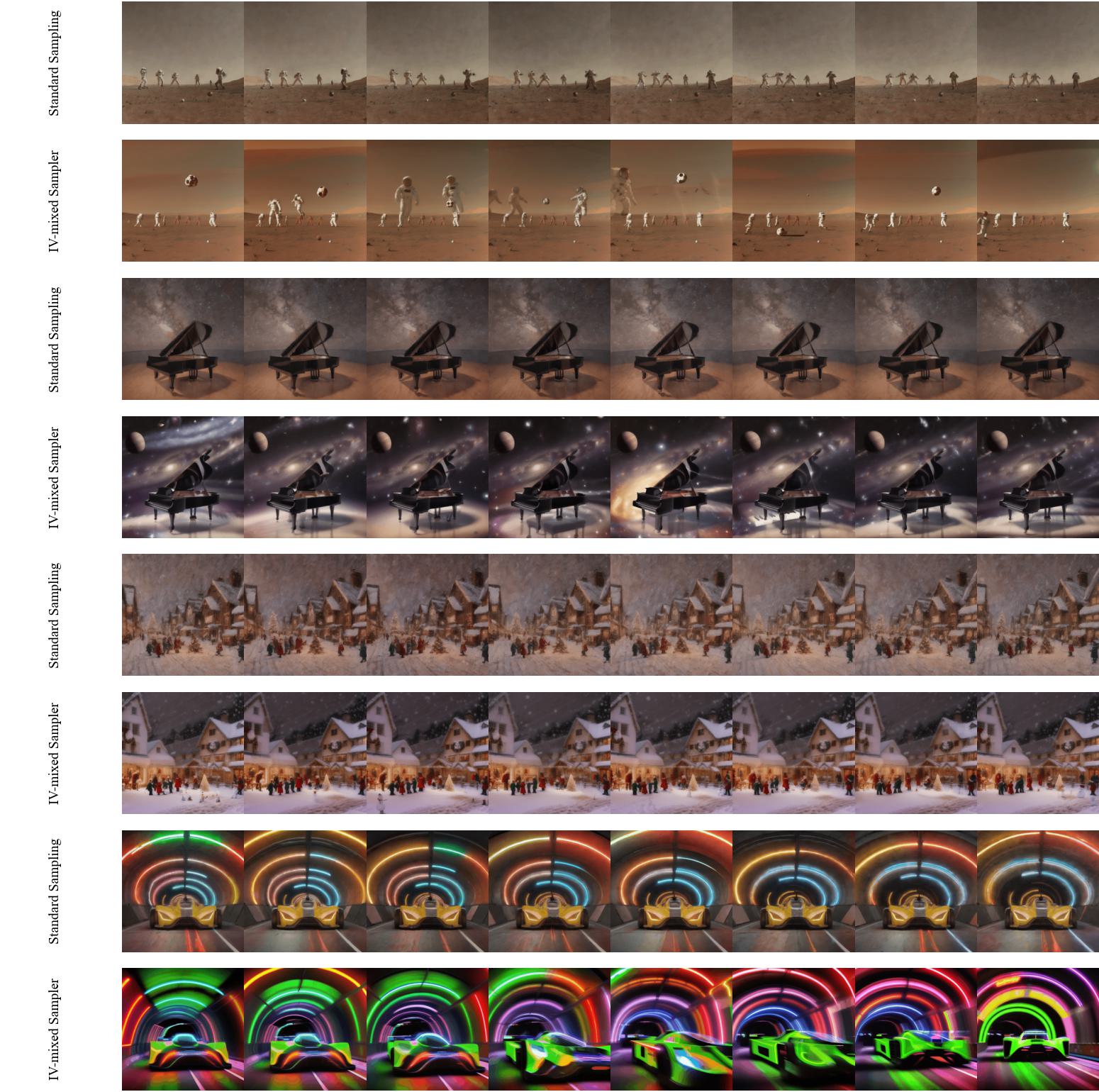}
\vspace{-7pt}
\caption{The synthesized video visualization of {Animatediff (SD V1.5, Motion Adapter V3)}.}
\vspace{-1.1ex}
\label{figure:animatediff_1}
\end{figure}

\begin{figure}[h]
\centering
\includegraphics[width=1.\textwidth,trim={0cm 0cm 0cm 0cm},clip]{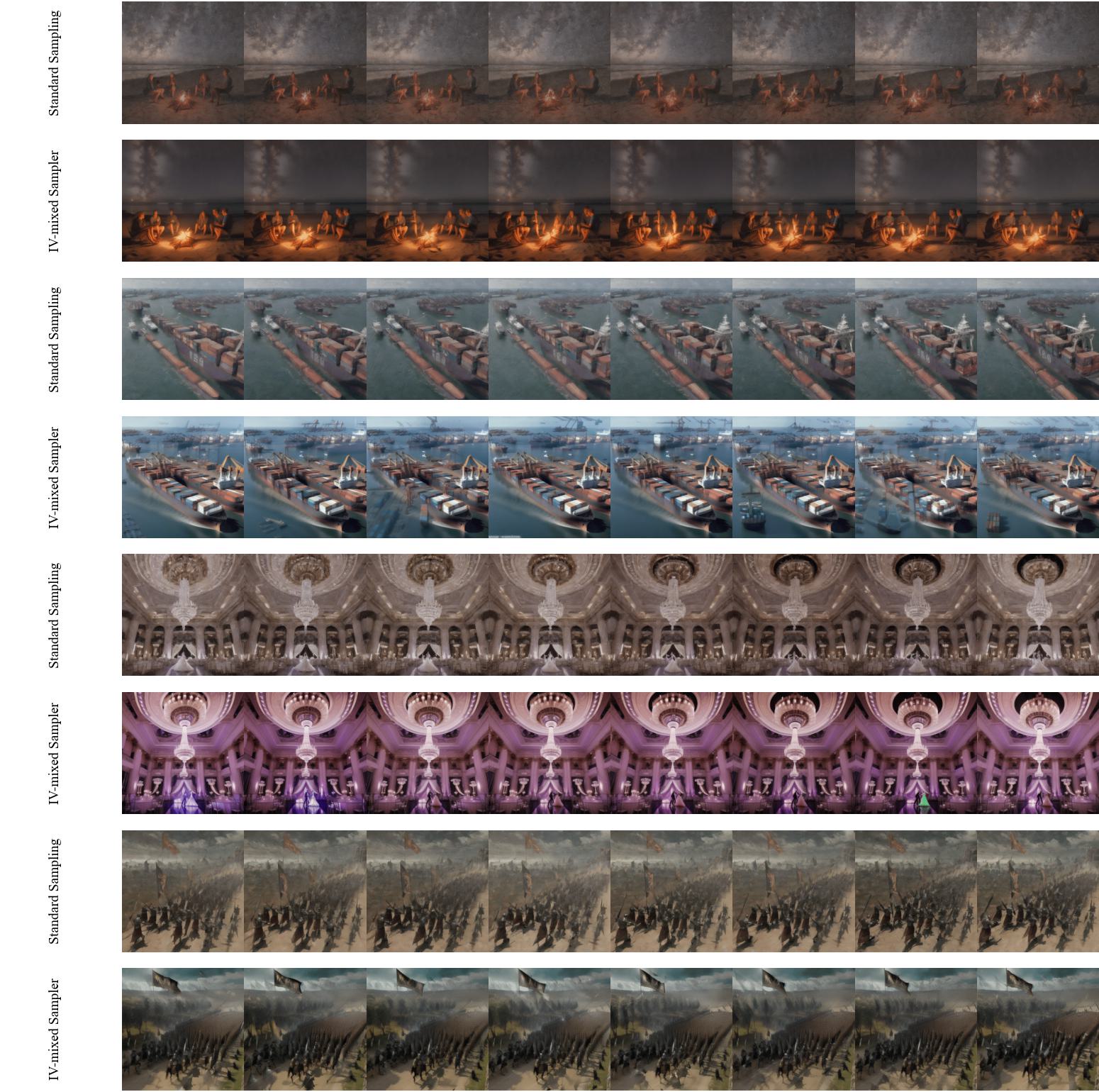}
\vspace{-7pt}
\caption{The synthesized video visualization of {Animatediff (SD V1.5, Motion Adapter V3)}.}
\vspace{-1.1ex}
\label{figure:animatediff_2}
\end{figure}

\begin{figure}[h]
\centering
\includegraphics[width=1.\textwidth,trim={0cm 0cm 0cm 0cm},clip]{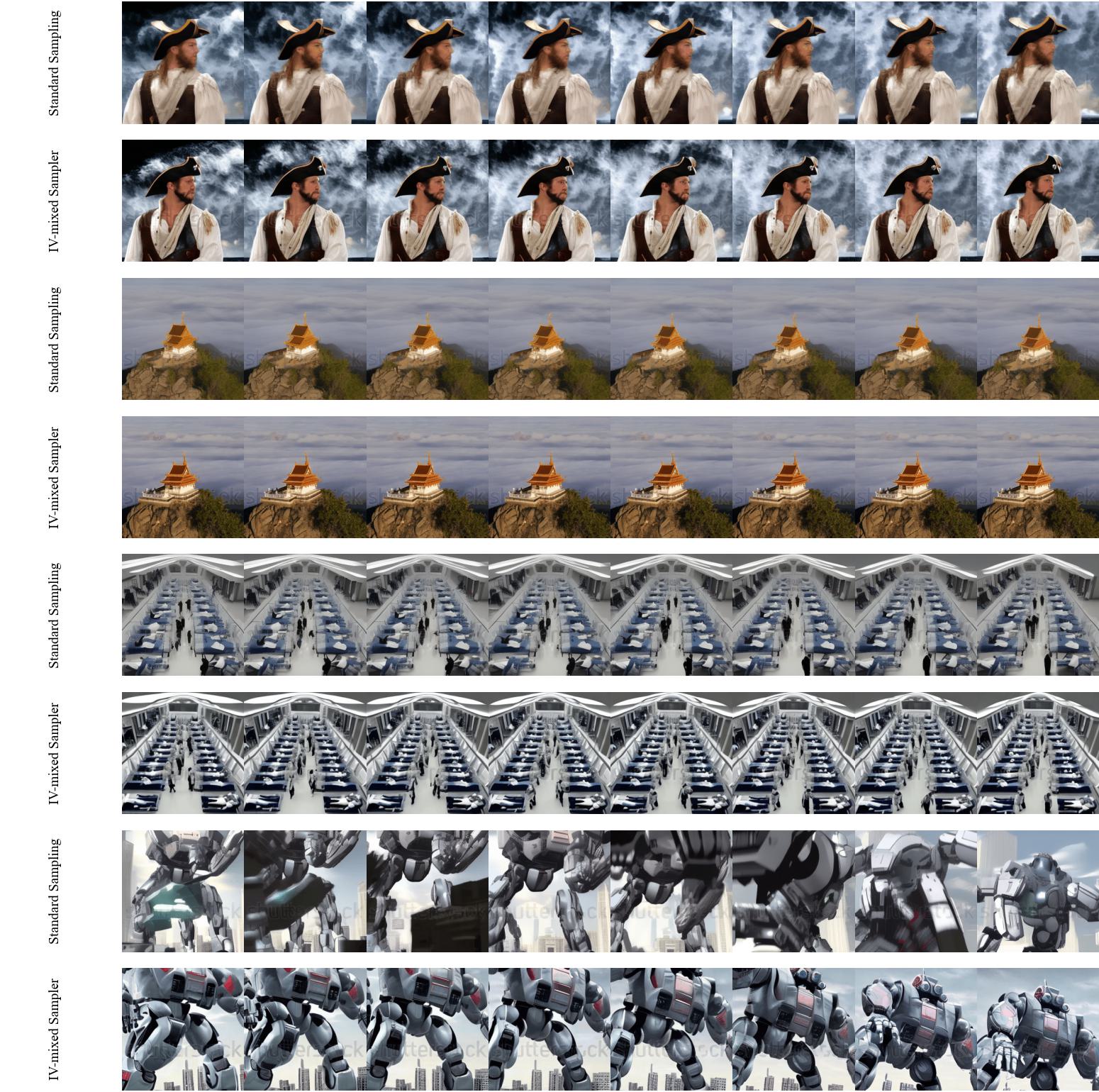}
\vspace{-7pt}
\caption{The synthesized video visualization of {ModelScope-T2V}.}
\vspace{-1.1ex}
\label{figure:modelscope}
\end{figure}

\begin{figure}[h]
\centering
\includegraphics[width=1.\textwidth,trim={0cm 0cm 0cm 0cm},clip]{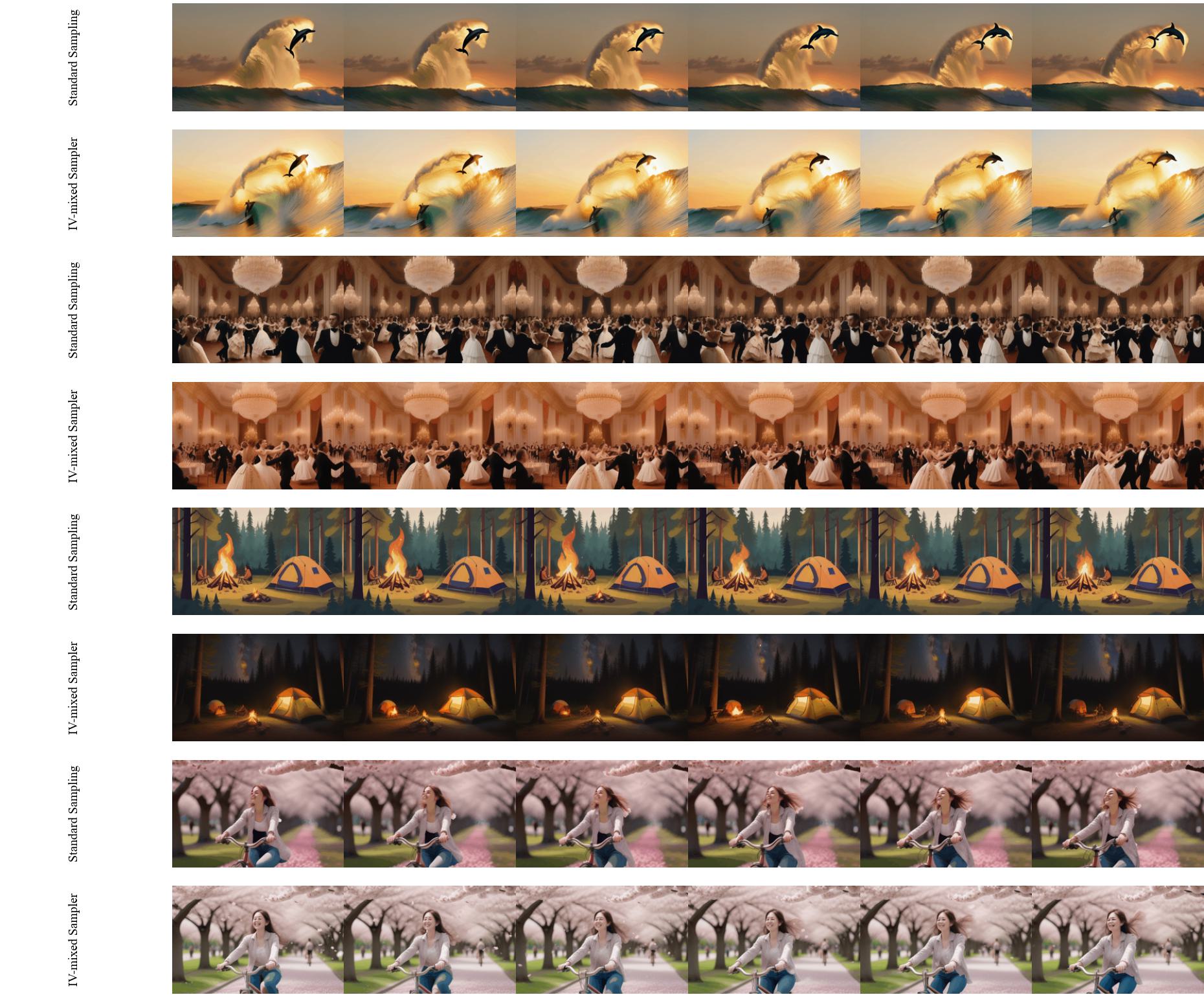}
\vspace{-7pt}
\caption{The synthesized video visualization of {VideoCrafterV2}.}
\vspace{-1.1ex}
\label{figure:videocrafterv2}
\end{figure}

In order to avoid the size of the paper being too large for the reader, we downsample the video frames and present them here. We present the synthesized video visualization of {Animatediff (SD V1.5, Motion Adapter V3)} in Fig.~\ref{figure:animatediff_1}-\ref{figure:animatediff_2}, the synthesized video visualization of {ModelScope-T2V} in Fig.~\ref{figure:modelscope} and the synthesized video visualization of {VideoCrafterV2} in Fig.~\ref{figure:videocrafterv2}.

\end{document}

%% file: math_commands.tex
\usepackage{amsmath,amsfonts,bm}

\def\eqref#1{equation~\ref{#1}}

\def\1{\bm{1}}

\DeclareMathAlphabet{\mathsfit}{\encodingdefault}{\sfdefault}{m}{sl}
\SetMathAlphabet{\mathsfit}{bold}{\encodingdefault}{\sfdefault}{bx}{n}



%% file: tables/chronomagic_150.tex
\begin{table*}[!t]
\centering
\vskip -0.01in
\small
\caption{\small \textbf{Quantitative comparison with popular heavy-inference algorithms on Chronomagic-Bench-150.} \Add{$[\cdot]$} stands for the improvement of \IVS\ compared to the ``Standard DDIM''. I4VGen is not compared across other architectures because its official implementation is limited to Animatediff.}
\scalebox{0.78}{
\begin{tabular}{clcccccc}
\toprule
{{Model}} & {{Method}} & Extra Model & {UMT-FVD (↓)} & {UMTScore (↑)} & {GPT4o-MTScore (↑)} \\
\midrule
\multirow{2}{*}{{VideoCrafterV2}} & \makecell{Standard DPM-Solver++} & \RR{\xmark} & 214.06 & 2.76 & 2.87 \\
& {\IVS\ (ours)} & \GG{\cmark} & \CC 210.57 \Add{3.49} & \CC 3.29 \Add{0.53} & \CC 3.30 \Add{0.43} \\
\midrule
\multirow{3}{*}{{ModelScope-T2V}} & {Standard DDIM} & \RR{\xmark} & 241.61 & 2.66 & 2.96 \\
& {FreeInit} & \RR{\xmark} & \CC 220.96 & \CB 3.01 & \CB 3.09 \\
& {\IVS\ (ours)} & \GG{\cmark} & \CB 234.90 \Add{6.71} & \CC 3.02 \Add{0.36} & \CC 3.14 \Add{0.18} \\
\midrule
\multirow{4}{*}{\makecell{{Animatediff}\\{(SD V1.5,}\\{Motion Adapter V3)}}} 
& {Standard DDIM} & \RR{\xmark} & 275.18 & \CB 2.82 & 2.83 \\
& {FreeInit} & \RR{\xmark} & 268.31 & \CB 2.82 & 2.59 \\
& {I4VGen} & \GG{\cmark} & \CC 227.21 & 2.66 & \CB 3.02 \\
& {\IVS\ (ours)} & \GG{\cmark} & \CB 228.59 \Add{46.59} & \CC 3.30 \Add{0.48}  & \CC 3.55 \Add{0.72} \\
\bottomrule
\end{tabular}
}
\label{tab:comparison}
\vskip -0.07in
\end{table*}

%% file: tables/chronomagic_1649.tex
\begin{table*}[!t]
\centering
\vskip -0.01in
\small
\caption{\small \textbf{Quantitative comparison with popular heavy-inference algorithms, including FreeInit and I4VGen, on Chronomagic-Bench-1649~\citep{yuan2024chronomagic}.} Yellow \colorbox{aliceblue!25}{cells} stands for the winner.}
\scalebox{0.78}{
\begin{tabular}{clccccc}
\toprule
{{Model}} & {{Method}} & {Extra Model} & {UMT-FVD (↓)} & {UMTScore (↑)} & {GPT4o-MTScore (↑)} \\
\midrule
\multirow{2}{*}{{VideoCrafterV2}} 
& \makecell{Standard DPM-Solver++}& \RR{\xmark} & 178.45 & 2.75 & 2.68 \\
& {\IVS\ (ours)} & \GG{\cmark} & \CC 172.02 \Add{6.43} & \CC 3.35 \Add{0.60} & \CC 3.04 \Add{0.36} \\
\midrule
\multirow{2}{*}{{ModelScope-T2V}} 
& {Standard DDIM} & \RR{\xmark} & 199.52 & 2.99 & 3.17 \\
& {\IVS\ (ours)} & \GG{\cmark} &\CC 197.44 \Add{2.08} & \CC 3.06 \Add{0.07} & \CC 3.25 \Add{0.08} \\
\midrule
\multirow{4}{*}{\makecell{{Animatediff}\\{(SD V1.5,}\\{Motion Adapter V3)}}} 
& {Standard DDIM} & \RR{\xmark} & 219.29 & 3.08 & 2.62 \\
& {FreeInit} & \RR{\xmark} & 209.62 & 3.08 & 2.73 \\
& {I4VGen} & \GG{\cmark} & 206.21 & 3.21 & 3.07 \\
& {\IVS\ (ours)} & \GG{\cmark} & \CC 192.72 \Add{26.57} & \CC 3.40 \Add{0.32} & \CC 3.34 \Add{0.72} \\
\bottomrule
\end{tabular}
}
\label{tab:comparison_1649}
\vskip -0.07in
\end{table*}

%% file: tables/traditional_bench_ucf.tex
\begin{table*}[!t]
\centering
\vskip -0.01in
\small

\renewcommand\arraystretch{1.2}
\setlength{\tabcolsep}{10pt}
\caption{\small \textbf{Quantitative comparison on UCF-101 datasets using Animatediff and ModelScope-T2V.} \Add{$[\cdot]$} indicates the improvement of \IVS\ compared to the ``Standard DDIM''.}
\scalebox{0.78}{
\begin{tabular}{ccccc}
\hline
Model & Method & Extra Model  & \makecell{FVD (↓)\\(StyleGAN)} & \makecell{FVD (↓)\\(VideoGPT)} \\ \hline
\multirow{5}{*}{\makecell{Animatediff\\(SD V1.5,\\Motion Adapter V3)}}& Standard DDIM & \RR{\xmark} & 815.08 & 819.93 \\
& FreeInit & \RR{\xmark} & 805.33 & 807.04 \\
& Unictrl & \RR{\xmark} & 1859.13 & 1863.35 \\
& I4VGen & \GG{\cmark} & 803.25 & 805.77\\
& \IVS\ (ours) & \GG{\cmark} & \CC 800.45 \Add{14.63} & \CC 804.88 \Add{15.05}\\ \hline
\multirow{2}{*}{ModelScope-T2V}& Standard DDIM & \RR{\xmark} & 1492.17 & 1484.97 \\
& \IVS\ (ours) & \GG{\cmark} & \CC 841.64 \Add{650.53} & \CC 838.05 \Add{646.92}\\
\bottomrule
\end{tabular}
}
\label{tab:traditional_bench_animatediff}
\vskip -0.01in
\end{table*}

%% file: tables/traditional_bench_msr.tex
\begin{table*}[!t]
\centering
\vskip -0.01in
\small

\renewcommand\arraystretch{1.2}
\setlength{\tabcolsep}{10pt}
\caption{\small \textbf{Quantitative comparison on MSR-VTT using Animatediff and ModelScope-T2V.} \Add{$[\cdot]$} indicates the improvement of \IVS\ compared to the ``Standard DDIM''.}
\scalebox{0.78}{
\begin{tabular}{ccccc}
\hline
Model & Method & Extra Model & \makecell{FVD (↓)\\(StyleGAN)} & \makecell{FVD (↓)\\(VideoGPT)} \\ \hline
\multirow{4}{*}{\makecell{Animatediff \\ (SD V1.5,\\Motion Adapter V3)}}& Standard DDIM & \RR{\xmark} & 762.23 & 761.01 \\
& FreeInit & \RR{\xmark} & 732.57 & 731.04 \\
& I4VGen & \GG{\cmark} & 741.56 & 740.13 \\
& \IVS\ (ours) & \GG{\cmark} & \CC 721.32 \Add{40.91}& \CC 719.9 \Add{41.11}\\ \hline
\multirow{2}{*}{ModelScope-T2V}& Standard DDIM & \RR{\xmark} & 739.12 & 737.24 \\
& \IVS\ (ours) & \GG{\cmark} & \CC 603.31 \Add{135.81}& \CC 601.89 \Add{135.35}\\
\bottomrule
\end{tabular}
}
\label{tab:traditional_bench_modelscope}
\vskip -0.01in
\end{table*}

%% file: tables/appendix_chronomagic_150_modelscope.tex
\begin{table*}[!t]
\centering
\vskip -0.01in
\small
\caption{\small \textbf{Ablation studies with Modelscope-T2V on Chronomagic-Bench-150.}}
\scalebox{0.78}{
\begin{tabular}{clcccccc}
\toprule
{{Model}} & {{Method}} & Extra Model & {UMT-FVD (↓)} & {UMTScore (↑)} & {GPT4o-MTScore (↑)} \\
\midrule
\multirow{3}{*}{{ModelScope-T2V}} & {Standard DDIM} & \RR{\xmark} & 241.61 & 2.66 & 2.96 \\
& {\IVS\ (Mini SD)} & \GG{\cmark} & 247.99 & 2.63 & 3.00 \\
& {\IVS\ (SD V1.5)} & \GG{\cmark} & \CB 234.90 \Add{6.71} & \CC 3.02 \Add{0.36} & \CC 3.14 \Add{0.18} \\
\bottomrule
\end{tabular}
}
\label{tab:comparison_modelscope_appendix}
\vskip -0.01in
\end{table*}

%% file: tables/chronomagic_150_videocrafterv2_appendix.tex
\begin{table*}[!t]
\centering
\vskip -0.01in
\small
\caption{\small \textbf{Performance comparison of VideoCrafterV2 across different $z\%$ settings.}}
\scalebox{0.86}{
\begin{tabular}{clccc}
\toprule
{{Model}} & {$z\%$} & {UMT-FVD (↓)} & {UMTScore (↑)} & {GPT4o-MTScore (↑)} \\
\midrule
\multirow{5}{*}{{VideoCrafterV2}} 
& Standard DDIM & 214.06 & 2.76 & 2.87 \\
& {33.3\%} & 212.74 & 3.08 & 3.02 \\
& {50.0\%} & \CC 208.67 \Add{5.39} & 3.23 & 3.18 \\
& {66.7\%} & 210.57 & \CC 3.29 \Add{0.53} & \CC 3.30 \Add{0.43} \\
& {75.0\%} & 211.87 & 3.28 & 3.28 \\
\bottomrule
\end{tabular}
}
\label{tab:comparison_z_percent}
\vskip -0.01in
\end{table*}

%% file: algorithms/iviv_sampler.tex
\begin{algorithm}[t]
    \caption{Pseudo code of \textsc{\IVS\ (\textit{w.r.t.}, ``IV-IV'')} (\(\imineq{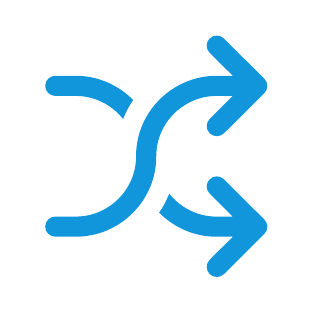}{2.4}\)) in a PyTorch-like style.}
    \label{algo:ctc}
    \footnotesize
    \begin{alltt}
    \color{ForestGreen}\tiny
# x_T: the initial noise
# N: the number of sampling step
# scheduler: a scheduler similar to that of diffusers
# inv_scheduler: an inversion scheduler similar to that of diffusers
# F_idm, F_vdm: the backbones of IDM and VDM
# pe_idm, pe_vdm: the prompt embedding of IDM and VDM
# w_idm_begin, w_idm_end, w_vdm_begin, w_vdm_end, rho: \/
# \quad the hyperparameter of the dynamic CFG scale
\end{alltt}
\vspace{-6pt}
\begin{alltt}\tiny
def get_sigmas_karras(n, sigma_begin, sigma_end, rho=7.0):
        ramp = torch.linspace(0, 1, n)
        begin_inv_rho = sigma_begin ** (1 / rho)
        end_inv_rho = sigma_end ** (1 / rho)
        sigmas = (begin_inv_rho +
        \quad ramp * (end_inv_rho - begin_inv_rho)) ** rho
        return sigmas \color{ForestGreen}# Constructs the noise schedule of Karras et al. \color{Black}

def sampling(latent, scheduler, inv_scheduler, F_idm, F_vdm, **kwargs):
        pe_idm, pe_vdm = kwargs.get("pe_idm", None), kwargs.get("pe_vdm", None)
        w_idm_begin, w_idm_end, w_vdm_begin, w_vdm_end, rho =
        \quad kwargs.get("w_idm_begin", 4), kwargs.get("w_idm_end", 4),
        \quad kwargs.get("w_vdm_begin", 4), kwargs.get("w_vdm_end", 4),
        \quad kwargs.get("rho", 1)
        N = kwargs.get("N", 50)
        idm_CFG = get_sigmas_karras(N, w_idm_begin, w_idm_end, rho) \color{ForestGreen}# Dynamic CFG scale \color{Black}
        vdm_CFG = get_sigmas_karras(N, w_vdm_begin, w_vdm_end, rho)
        
        for step in range(N):
            t = scheduler.timesteps[step]
            p_t = min(t + scheduler.config.num_train_timesteps
            \quad // scheduler.num_inference_steps, 
            \quad scheduler.config.num_train_timesteps-1)
            pp_t = min(t + 2 * scheduler.config.num_train_timesteps
            \quad // scheduler.num_inference_steps, 
            \quad scheduler.config.num_train_timesteps-1)
            \color{ForestGreen}# Perform sampling with IDM \color{Black}
            latent = einops.rearrange(latent, "b c t h w -> (b t) c h w")
            n_latent =  torch.cat([latent] * 2)
            n_latent = scheduler.scale_model_input(n_latent, t)
            n_pred = F_idm(n_latent, t, pe_idm)
            n_pred_un, n_pred_te = n_pred.chunk(2)
            n_pred = n_pred_un + idm_CFG[step] * (n_pred_te - n_pred_un)
            latent = inv_scheduler.step(n_pred, p_t, latent)
            latent = einops.rearrange(latent, "(b t) c h w -> b c t h w", t=(...) )
            \color{ForestGreen}# Perform sampling with VDM \color{Black}
            n_latent =  torch.cat([latent] * 2)
            n_latent = scheduler.scale_model_input(n_latent, p_t)
            n_pred = F_vdm(n_latent, p_t, pe_vdm)
            n_pred_un, n_pred_te = n_pred.chunk(2)
            n_pred = n_pred_un + vdm_CFG[step] * (n_pred_te - n_pred_un)
            latent = inv_scheduler.step(n_pred, pp_t, latent)
            \color{ForestGreen}# Perform sampling with IDM \color{Black}
            ...
            n_latent = scheduler.scale_model_input(n_latent, pp_t)
            ...
            latent = scheduler.step(n_pred, pp_t, latent)
            \color{ForestGreen}# Perform sampling with VDM \color{Black}
            ...
            n_latent = scheduler.scale_model_input(n_latent, p_t)
            ...
            latent = scheduler.step(n_pred, p_t, latent)
            \color{ForestGreen}# Perform sampling with VDM \color{Black}
            ...
            n_latent = scheduler.scale_model_input(n_latent, t)
            ...
            latent = scheduler.step(n_pred, t, latent)
        return decode_latent(latent)             \color{ForestGreen}# transfer latent to video \color{Black}
\color{ForestGreen}# IV-mixed Sampler \color{Black}
video = sampling(x_T, scheduler, inv_scheduler, ... )
\end{alltt}
\end{algorithm}